\documentclass[lettersize,journal]{IEEEtran}
\usepackage{algorithmicx,algorithm}
\usepackage{array}
\usepackage[caption=false,font=normalsize,labelfont=sf,textfont=sf]{subfig}
\usepackage{textcomp}
\usepackage{stfloats}
\usepackage{url}
\usepackage{verbatim}
\usepackage{graphicx}
\usepackage{cite}
\usepackage{mathrsfs}
\usepackage{amsmath,amssymb,amsfonts}
\usepackage{graphicx}
\usepackage{booktabs} 
\usepackage{multirow}
\usepackage{amsthm}
\usepackage[noend]{algpseudocode}
\usepackage{tabularx}
\begin{document}

\title{From Incomplete Coarse-Grained to Complete Fine-Grained:
A Two-Stage Framework for Spatiotemporal Data Reconstruction}

\author{Ziyu~Sun, Haoyang~Su, En~Wang, Funing~Yang, Yongjian~Yang, Wenbin~Liu
\thanks{\IEEEcompsocthanksitem Ziyu~Sun, Haoyang~Su, En~Wang, Funing~Yang, Yongjian~Yang, Wenbin~Liu are with the College of Computer Science and Technology, Jilin University, Changchun 130012, China, and also with the Key Laboratory of Symbolic Computation and Knowledge Engineering of Ministry of Education, Jilin University, Changchun 130012, China
(e-mail: sunzy2121@mails.jlu.edu.cn ;suhy2121@mails.jlu.edu.cn; wangen@jlu.edu.cn; yfn@jlu.edu.cn; yyj@jlu.edu.cn; liuwenbin@jlu.edu.cn).
}
\thanks{(Ziyu Sun and Haoyang Su contributed equally to this work.) (Corresponding author: Wenbin~Liu.)}}



\maketitle

\begin{abstract}
With the rapid development of various sensing devices, spatiotemporal data is becoming increasingly important nowadays. However, due to sensing costs and privacy concerns, the collected data is often incomplete and coarse-grained, limiting its application to specific tasks. To address this, we propose a new task called spatiotemporal data reconstruction, which aims to infer complete and fine-grained data from sparse and coarse-grained observations. To achieve this, we introduce a two-stage data inference framework, DiffRecon, grounded in the Denoising Diffusion Probabilistic Model (DDPM). In the first stage, we present Diffusion-C, a diffusion model augmented by ST-PointFormer, a powerful encoder designed to leverage the spatial correlations between sparse data points. Following this, the second stage introduces Diffusion-F, which incorporates the proposed T-PatternNet to capture the temporal pattern within sequential data. Together, these two stages form an end-to-end framework capable of inferring complete, fine-grained data from incomplete and coarse-grained observations. We conducted experiments on multiple real-world datasets to demonstrate the superiority of our method.
\end{abstract}

\begin{IEEEkeywords}
Spatiotemporal Data, Data Completion, Fine-grained Inference.
\end{IEEEkeywords}

\section{Introduction}
\IEEEPARstart{S}{patiotemporal} data is of paramount importance in numerous applications, ranging from environmental monitoring to urban planning and beyond \cite{han2021joint,yuan2021survey,liu2022practical,cheng2020short,meng2021cross}. Existing methods for collecting spatiotemporal data primarily rely on either fixed sensors or mobile sensing technologies\cite{wang2020survey}. However, these approaches face significant limitations in terms of cost and complexity, resulting in that we can only obtain the coarse-grained spatiotemporal data, and it often fails to cover all spatiotemporal areas comprehensively \cite{ouyang2020fine,liu2023pristi}. Obviously, such coarse and sparse data is insufficient to support various urban computing tasks. This limitation raises an intuitive and important new problem called spatiotemporal data reconstruction, i.e., how can we infer the complete and fine-grained spatiotemporal data from the sparse and coarse-grained observations?  


Currently, there are efforts to adapt super-resolution methods from computer vision to infer fine-grained results from coarse-grained observation. \cite{liang2019urbanfm} employs distributional upsampling to obtain fine-grained results, while \cite{xu2023diffusion} investigates generative models to achieve similar outcomes. However, these methods are inadequate for handling sparse data, which is common in practical data collection. It is obvious that spatiotemporal reconstruction is significantly more challenging than the super-resolution of spatiotemporal data, with the primary difficulty in capturing correlations within sparse data. Super-resolution requires understanding spatial structural information from coarse-grained data to generate fine-grained details. However, missing data disrupts the spatial structural information, thereby affecting our understanding of spatial relationships. For example, if we don't know there is a hospital on the map, we cannot infer the fine-grained structure of the hospital. Therefore, how to understand the complicated spatial correlation from only sparse observation is the first challenge.

\begin{figure}[!t]
\centerline{\includegraphics[width=1.0\linewidth]{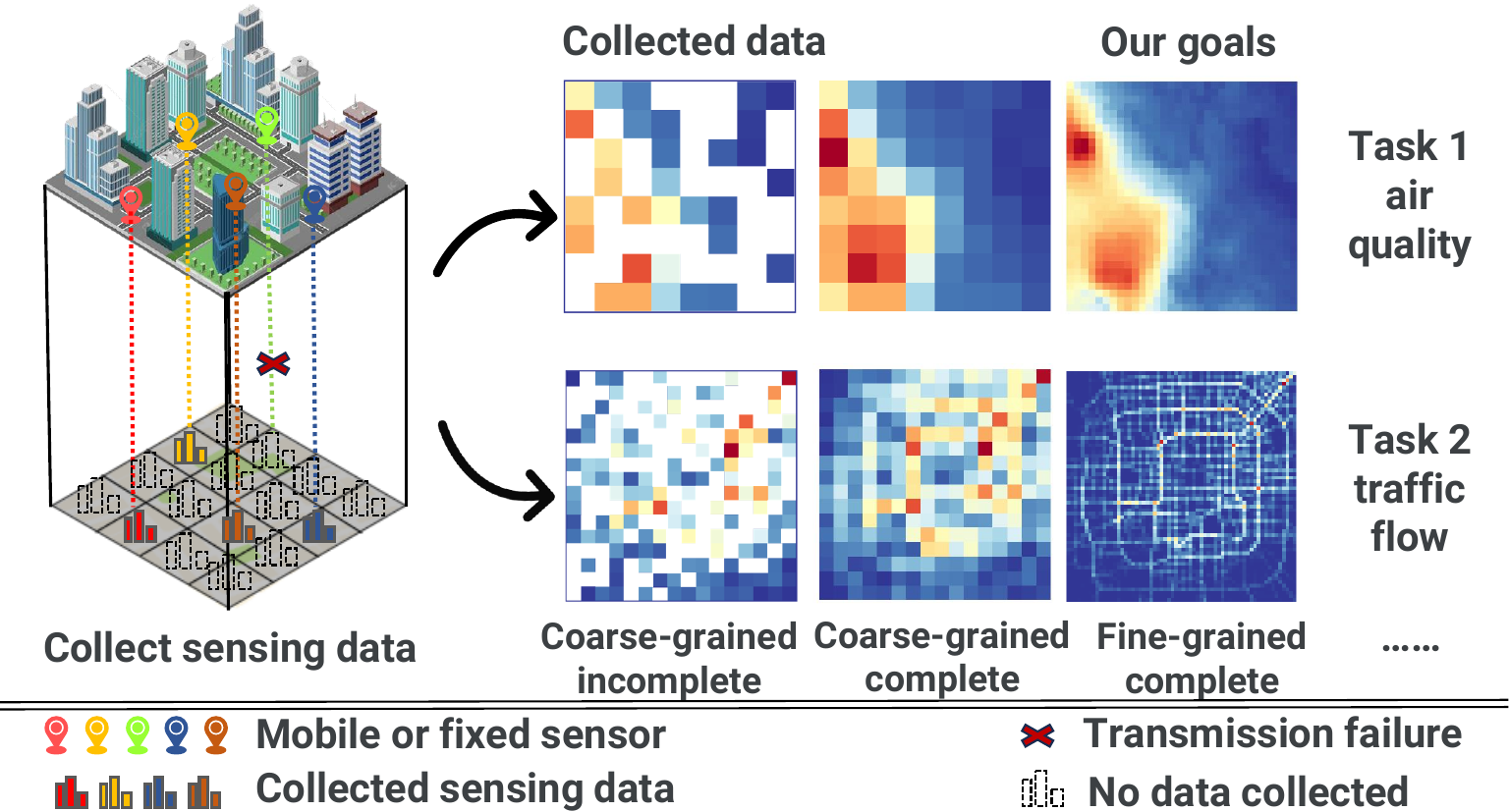}}
\caption{Task definition of spatiotemporal data reconstruction. The blank subregion indicates that no data was collected. }
\label{fig:task}
\vspace{-15pt}
\end{figure}

Even when spatial correlations can be gleaned from incomplete observations, temporal correlations remain challenging for existing work \cite{liang2019urbanfm, zhou2020enhancing} to handle. Spatiotemporal data typically exhibits strong periodicity and trends over time \cite{wang2020calendar}, which are characteristics not present in image or video data used in computer vision. Moreover, these temporal patterns exhibit complex cycles, such as daily, weekly, and monthly fluctuations, and are further associated with external factors like holiday setting and weather change. The challenge is compounded when observations are sparse and coarse-grained. Thus, our second challenge is to understand and utilize these temporal correlations to guild reconstruction.

Moreover, the issue of varying sparse data patterns adds another layer of complexity to the reconstruction process. Some studies in the traffic domain have explored tasks similar to spatiotemporal data reconstruction, but they rely on stationary and evenly distributed sensors \cite{wang2023urbanSTA}. However, sensor placement has an unexpectedly significant impact on the reconstruction outcome. The spatial distribution of sensed data may be uneven, and it can change over time, which has been shown to greatly affect the performance of existing methods. Thus, the third challenge is designing a general and robust method capable of handling data with varying sparse patterns. 

To address these challenges, we propose a novel two-stage model called DiffRecon which aims to effectively handle the sparse nature of the data, accurately capture spatiotemporal correlations, and account for varying sparse patterns, thereby providing a comprehensive solution to spatiotemporal data reconstruction. In the first stage, our primary goal is to enable the model to grasp the underlying structural information from the sparse observed data and augment it by filling in the missing data. We call this stage coarse-grained completion. By treating each observed data point as an individual spatiotemporal point and considering the relationships between them, we no longer attempt to learn the distribution of specific data pattern, making the approach robust to all sparse patterns. After coarse-grained completion, we infer fine-grained data based on the coarse-grained intermediate result. Temporal patterns such as periodicity and trend are considered in this stage, thanks to the preprocessing and augmentation performed in the coarse-grained completion stage. Our model will be jointly trained in two stages to provide end-to-end reconstruction results.

Technically, the two sub-models, namely Diffusion-C and Diffusion-F, both utilize the Denoising Diffusion Probabilistic Model (DDPM) \cite{ho2020denoising} as their foundational framework. DDPM has become a powerful model for data inference by artificially adding noise and training a denoising netework to remove it. Therefore, the key lies in the design of the denoising network. Typically, a denoising network follows an encoder-decoder structure. The encoder in traditional denoising networks focus solely on intra-map information within a single map, neglecting the rich inter-map information across the sequence. To address this, we augument the denoising network by designing task-specified encoders for each stage to enhance the generation performance. In Diffusion-C, the input spatiotemporal map sequences are incomplete, but the data granularity remains consistent throughout. This consistency allows us to leverage the relationships between spatiotemporal points (ST-points) to infer missing data points. Specifically, we propose ST-PointFormer, which calculates the relations between ST-points using a multi-head self-attention mechanism. In this process external features are also encoded to form the ST-points. Since ST-PointFormer considers the correlations between any two ST-points rather than memorizing specific data distributions, it is robust to varying sparse patterns.

In the second stage, the data granularity changes, but the input data maps are complete and thus the historical sequence at each spatial location can be obtained. For this reason, we design a temporal encoder called T-PatternNet, which explicitly models temporal patterns—such as periodicity—within the data sequence of each position. T-PatternNet is a novel method for modeling time sequence that excels at extracting multi-periodicity by using a Fast Fourier Transformer (FFT) along with 2D convolutions to analyze sequence data. The extracted temporal information is then aligned with the encoding from the traditional denoising network's encoder and pass the decoder together to acquire the denoised results.

To conclude, our main contributions are as follows:
\begin{itemize}
\item We propose a novel task called spatiotemporal data reconstruction. Spatiotemporal data reconstruction aims to infer complete fine-grained spatiotemporal map from sparse coarse-grained observed data.  Additionally, no specific sparse pattern of observed data is required in this task.
\item We propose a two-stage method called DiffRecon as a solution to spatiotemporal data reconstruction. With task-specialized denoising network, we successfully apply the diffusion method on spatiotemporal data, fully considering the spatiotemporal characteristics of observed data. 
\item We conduct extensive experiments on datasets exhibiting three representative sparse patterns to evaluate the effectiveness of our method. We also conduct multiple ablation studies to demonstrate the contribution of each module.
\end{itemize}

The remainder of the paper is organized as follows. We introduce the related work in Section ~\ref{Related Work}, followed by a detailed problem formulation of spatiotemporal data reconstruction task in section ~\ref{Definitions and Problem Formulation}. After that, we introduce our proposed solution DiffRecon in section ~\ref{METHODOLOGY}. The experimental results are finally demonstrated in section ~\ref{EXPERIMENTS}.

\section{Related Work}
\label{Related Work}

\begin{figure*}[!t]
\centering
\includegraphics[width=1.0\linewidth]{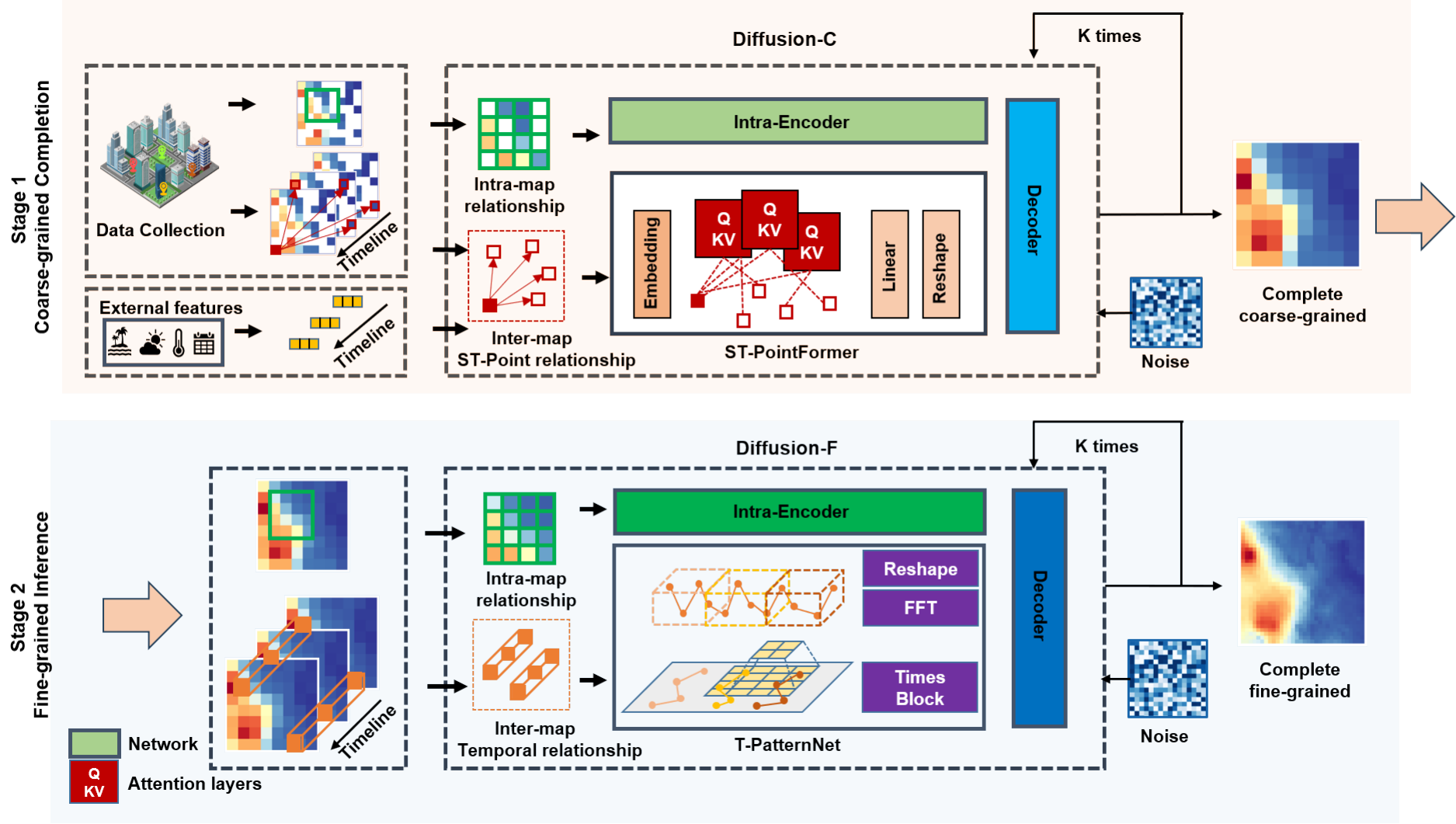}
\captionsetup{justification=centering}
\caption{The two-stage framework of DiffRecon.}
\label{fig:framework}
\vspace{-10pt}
\end{figure*}

\subsection{Spatiotemporal Data Completion}
Due to cost constraints and the prevalence of transmission errors in devices, the spatiotemporal data collected is often incomplete or even sparse. Spatiotemporal data completion methods aim to infer the complete data distribution from the incomplete data collected. In the early stages, researchers considered treating spatiotemporal data distributions divided into grid regions as images. For example, \cite{pathak2016context} designed a context encoder that understands the entire content of the image and proposes reasonable hypotheses for the missing parts. \cite{yu2018generative} proposed a feed-forward generative network that utilizes a novel contextual attention layer to exploit long-range spatial correlations. However, since these methods were designed for image restoration, their application to spatiotemporal data completion is less effective due to the lack of consideration for the unique characteristics of spatiotemporal data.

In recent years, some algorithms designed for spatiotemporal data completion have emerged. \cite{wu2021spatial} proposed a Spatial Aggregation and Temporal Convolution Network that uses various aggregation functions to leverage spatial correlations and temporal convolution to capture temporal dependencies. \cite{xu2019fine} fused multiple sources of data and used tensor decomposition for fine-grained air quality inference. \cite{wang2020deep} considered the low-rank nature of spatiotemporal data matrices, combining matrix factorization with neural networks to propose Deep Matrix Factorization (DMF) for inferring complete data. Based on Transformer, \cite{wang2023spatiotemporal} proposed a novel framework for spatiotemporal data completion and prediction. However, these methods perform spatiotemporal data completion at the same granularity, without considering the transformation relationships between different granularities of data.

\subsection{Spatiotemporal Data Super-Resolution}
he complexity and accuracy requirements of real-world tasks increase, the demand for fine-grained spatiotemporal data also grows, making spatiotemporal data super-resolution a feasible solution. It aims to infer fine-grained spatiotemporal data distribution from coarse-grained spatiotemporal data. In the early stages, methods for image super-resolution were applied to spatiotemporal super-resolution tasks \cite{wang2020survey}. For example, SRCNN \cite{dong2015image} was the first to combine convolutional neural networks with bicubic interpolation methods to achieve an end-to-end super-resolution algorithm. SRResNet \cite{ledig2017photo} adopted deep convolutional networks and residual learning strategies to achieve better super-resolution results. However, these algorithms were designed for image super-resolution rather than spatiotemporal data, failing to consider the unique characteristics of spatiotemporal data, such as spatiotemporal relationships and external factors.

UrbanFM \cite{liang2019urbanfm} designed an external factor fusion network to extract external features, such as weather and temperature, combining them with the inference network to achieve better results in spatiotemporal super-resolution tasks compared to traditional image super-resolution methods. Building on UrbanFM, UrbanPy \cite{ouyang2020fine} designed a pyramid architecture with multiple components, further improving the performance. FODE \cite{zhou2020enhancing} extends neural ODEs by introducing an affine coupling layer, allowing for more accurate and efficient spatial correlation estimation. STCF \cite{xu2023spatial} designed two spatiotemporal contrastive pre-training networks and a fine-tuning network that integrates learned features, achieving good inference results while mitigating the overfitting problem of historical methods. However, these methods often assume that the collected coarse-grained spatiotemporal data is complete.

Considering that the spatiotemporal data collected in real-world scenarios is often incomplete, \cite{li2022fine} designed a multi-task learning model called MT-CSR, which simultaneously considers data completion and super-resolution. \cite{zheng2023diffuflow} further addressed the impact of noise and data sparsity, but each time slice (with incomplete coarse-grained data) requires a large amount of complete fine-grained data from the past for assistance, making it difficult to apply in real-world scenarios. \cite{wang2023urbanSTA} designed a two-stage framework with spatiotemporal attention learning, improving space-time representations, but it can only handle cases where the missing positions are fixed, ignoring the diversity of sparse patterns in real-world scenarios. Therefore, we aim to propose a generalized model for inferring complete fine-grained spatiotemporal data from incomplete coarse-grained data to handle various situations.

\subsection{Denoising Diffusion Probabilistic Model}
In recent years, diffusion models have become the most popular generative models, especially excelling in image generation tasks in computer vision \cite{rombach2022high,croitoru2023diffusion}. Beyond that, diffusion models have also shown significant potential in various fields such as natural language processing \cite{li2022diffusion,gong2022diffuseq}, multimodal modeling \cite{gu2022vector,bao2023one}, time series data modeling \cite{tashiro2021csdi,kong2020diffwave}, reinforcement learning \cite{lidiffstitch,kang2024efficient}, and robust learning \cite{nie2022diffusion,sun2023enhance}. In this paper, we adopt diffusion models as our foundational framework, complemented by two auxiliary modules, to achieve fine-grained spatiotemporal data inference.

\section{Definitions and Problem Formulation}
\label{Definitions and Problem Formulation}
\paragraph*{\textbf{Definition 1. Value of Regions}}
Based on geographic location information (longitude and latitude), we divide a region into an $I\times J$ grid map. Let all sub-regions be denoted as $\textbf{R}=\{r_{1,1},r_{1,2},...,r_{m,n},...,r_{I,J}\}$, where $r_{m,n}$ represents the sub-region in the $m$-th row and $n$-th column of the grid map. The average of all values uploaded by users within each sub-region is taken as the overall value of the region.

\paragraph*{\textbf{Definition 2. Coarse-grained and Fine-grained Spatiotemporal Data Distribution Map}}
Given a magnification factor $N$, the coarse-grained grid map divides the geographic area into $I\times J$ sub-regions, while the fine-grained grid map divides the area into $NI\times NJ$ sub-regions. Based on the concept of region value in Definition 1, we obtain the values of each coarse and fine-grained sub-region under the two divisions, forming coarse and fine-grained spatiotemporal data distribution maps. We denote the real data of the sub-region in the $i$-th row and $j$-th column as $x_{i,j}$, the complete coarse-grained spatiotemporal distribution map as $\textbf{X}_{\text{comp,cg}}=\{x_{1,1},...,x_{i,j},...,x_{I,J}\}$, and the complete fine-grained spatiotemporal distribution map as $\textbf{X}_{\text{comp,fg}}=\{x_{1,1},...,x_{i,j},...,x_{NI,NJ}\}$, where \textit{comp} represents complete, \textit{cg} represents coarse-grained, and \textit{fg} represents fine-grained.

\paragraph*{\textbf{Definition 3. Incomplete Coarse-grained Spatiotemporal Data Distribution Map}}
Due to objective factors such as transmission errors in user-uploaded data, some users' mobile devices only being able to collect coarse-grained data, and subjective reasons such as reducing the number of recruited users to lower costs, the obtained coarse-grained data distribution map is often incomplete. We use a binary variable $c_{i,j}$ to record whether data has been collected for a sub-region: if so, then $c_{i,j}=1$; otherwise, $c_{i,j}=0$. We obtain $x^{'}_{i,j}=x_{i,j}\times c_{i,j}$ and denote $\textbf{C}=\{c_{1,1},...,c_{i,j},...,c_{I,J}\}$. We denote $\textbf{X}_{\text{inc,cg}}=\{x^{'}_{1,1},...,x^{'}_{i,j},...,x^{'}_{I,J}\}$ as the collected coarse-grained sensing data, where \textit{inc} indicates that the collected data is incomplete. Therefore, We have
\begin{align}
\textbf{X}_{\text{inc,cg}}=\textbf{X}_{\text{comp,cg}}\odot \textbf{C},
\end{align}
where $\odot$ represents the Hadamard product. Note that $c_{i,j}$ can remain constant or vary across all timestamps, and the $0/1$ values may be distributed evenly or unevenly within each $c_{i,j}$. We refer to these different configurations of $\textbf{C}$ as 'sparse patterns,' as they reflect the varying degrees of data sparsity.

\paragraph*{\textbf{Problem}}
Given a magnification factor $N$ and a set of incomplete coarse-grained spatiotemporal data distribution maps $\{\textbf{X}^1_{\text{inc,cg}},\textbf{X}^2_{\text{inc,cg}},...,\textbf{X}^t_{\text{inc,cg}}\}\in {\mathbb{R}}^{I\times J}$, our goal is to infer the complete fine-grained data distribution maps $\{\textbf{X}^1_{\text{comp,fg}},\textbf{X}^2_{\text{comp,fg}},...,\textbf{X}^t_{\text{comp,fg}}\}\in {\mathbb{R}}^{NI\times NJ}$. It is worthy noting that the sparse pattern can be arbitrary in the task setting.

\section{METHODOLOGY}
\label{METHODOLOGY}
\subsection{Overall Structure}
In this section, we detail our model which consists of two main components named Diffusion-C and Diffusion-F for coarse-grained completion task and fine-grained inference task respectively. We employ the DDPM as the framework for both Diffusion-C and Diffusion-F to leverage its capacity to understand data distributions. Given the incomplete coarse-grained input $\textbf{X}^t_{inc, cg}$ and history series $\{\textbf{X}^i_{inc, cg}\}_{i = 0 \sim t-1}$, we first do data completion with Diffusion-C by extracting the spatial correlations within $\textbf{X}^t_{inc, cg}$ and the relations between $\{\textbf{X}^i_{inc, cg}\}_{i = 0 \sim t}$. After we obtain the complete coarse-grained results $\textbf{X}^t_{comp, cg}$ from the first stage, we use Diffusion-F to infer fine-grained data considering the spatial correlations in $\textbf{X}^t_{comp, cg}$ and the temporal patterns within the temporal series $\{\textbf{X}^i_{comp, cg}\}_{i = 0 \sim t}$. Diffusion-C and Diffusion-F are trained jointly to provide end-to-end inference.

Generally, DDPM learns the underlying distribution of given data by artificially adding noise and training a predictor to remove the noise. In our model, the predictor in each stage should utilize the spatiotemporal characteristics of the known data and fully leverage external domain knowledge to help the predictor accurately forecast noise. To achieve this goal, we design sub-task specialized encoders for each stage and utilize a decoder to predict the final noise and iteratively recover the reconstructed results from noise. We also incorporate external factors, such as weekly indices and holiday information, into the predictor through the use of embedding. For the two stages, due to the difference in input-output data and tasks, we utilize distinct encoding structures to tailor each stage to its specific task. We observe that the granularity remain consistent in the first stage, which allows us to calculate the similarity between the data points of each historical moment, providing clues for completing the missing values at the current time. This is achieved by utilizing self attention mechanism in the encoder called ST-PointFormer at the first stage. Conversely, data granularity changes at the second stage, yet there are no missing values in the input data, making it convenient to extract temporal patterns such as periodicity from the sequence. Therefore, inspired by \cite{wu2022timesnet}, we use a powerful TimesBlock as a feature extractor in the temporal encoder at the second stage to focus on associated temporal patterns.

In Section ~\ref{subsec:Conditional Diffusion}, we will present the framework of DDPM and the overall structures of our predictors.  In section ~\ref{ST-PointFormer} and section ~\ref{T-PatternNet},  we will introduce the predictor structure of the Diffusion-C and Diffusion-F in technical detail respectively, followed by the training strategy of our two-stage models which will be introduced in Section ~\ref{Training Strategy}.

\begin{figure*}[!t]
\centerline{\includegraphics[width=1.0\linewidth]{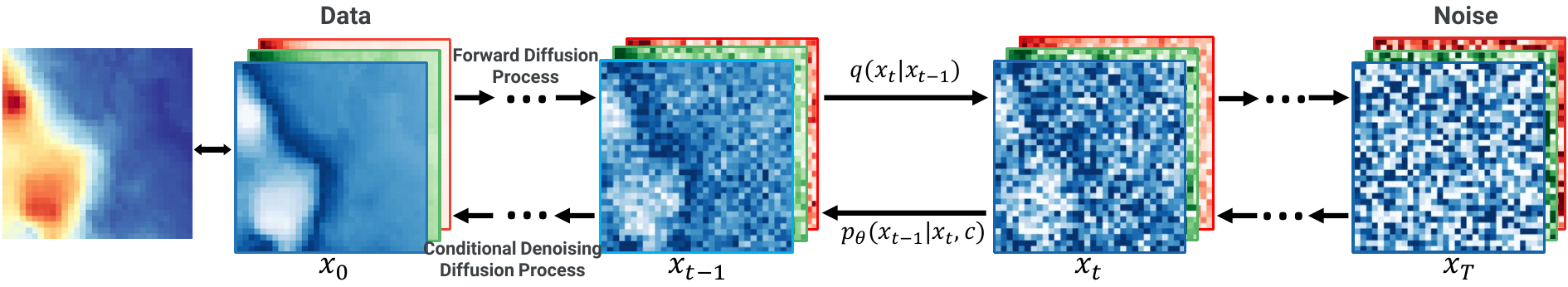}}
\caption{ Two Markov processes of Conditional Denoising Diffusion Probabilistic Model. }
\label{fig:diffusion}
\vspace{-15pt}
\end{figure*}

\subsection{Conditional Diffusion}
\label{subsec:Conditional Diffusion}
The Denoising Diffusion Probabilistic Model (DDPM) is an emerging generative model that learns the probability distribution of data through a process of adding noise and denoising, as shown in Figure~\ref{fig:diffusion}. It demonstrates impressive data generalization capabilities and noise resistance, while maintaining a stable training process.  Based on  that, Conditional DDPM further constrains the generated process by using conditional guidance and allows us to acquire desired outcomes based on our conditions. In both stages we utilize the conditional DDPM as the framework and work on improving its denoising process by designing task-specialized noise predictor.

\subsubsection{Forward Diffusion Process}
Intuitively, the forward diffusion process involves gradually adding Gaussian noise to a sample until the data becomes random noise. For data \(s_0\) sampled from the real data distribution \(q(s_0)\), each step of the total \(T\) steps in the diffusion process adds Gaussian noise to the data obtained from the previous step as follows:
\begin{align}
q(s_\tau|s_{\tau-1}) &= \mathcal{N}(s_\tau;\sqrt{1-\beta_\tau}s_{\tau-1},\beta_\tau I),
\end{align}
where $\{\beta_\tau\}^T_{\tau=1}$ represents the variance used at each step, ranging between 0 and 1. If the number of diffusion steps $T$ is sufficiently large, the final $x_T$ almost completely loses the characteristics of the original data distribution and becomes pure Gaussian noise.  Each step in the diffusion process generates a data $s_\tau$ with added noise, and the entire forward diffusion process forms a Markov chain:
\begin{align}
q(s_{1:T}|s_0) &= \prod_{\tau=1}^{T}q(s_\tau|s_{\tau-1}).
\end{align}
Given that the forward diffusion process is typically fixed, employing a pre-defined variance schedule, we can sample $s_\tau$ at any step \(\tau\) directly from the original data $s_0$ using the reparameterization trick. By defining $\alpha_\tau=1-\beta_\tau$ and $\overline{\alpha}_\tau=\prod_{i=1}^{\tau}\alpha_i$, it follows that:
\begin{align}
s_\tau &= \sqrt{\overline{\alpha}_\tau}x_0+\sqrt{1-\overline{\alpha}_\tau}\epsilon.
\end{align}
Consequently, we derive:
\begin{align}
q(s_\tau|s_0) &= \mathcal{N}(s_\tau;\sqrt{\overline{\alpha}_\tau}s_0,(1-\overline{\alpha}_\tau)I).
\end{align}

\subsubsection{Conditional Denoising Process}
The forward diffusion process adds noise to data, whereas the reverse process is a denoising process. We start from a random noise $s_T\sim\mathcal{N}(0,I)$ and train a noise predictor to progressively denoise it to enable sample generation. Unlike traditional unconditional generation, we need to utilize the observed data as conditions to guide the generation process during both the first and second stage. Specifically, we define the reverse process as a Markov chain composed of a series of Gaussian distributions parameterized by the noise predictor. We describe the individual step of the denoising diffusion process as:
\begin{align}
p_{\theta}(s_{\tau-1}|s_\tau,{\ell}) &= \mathcal{N}(s_{\tau-1};\mu_{\theta}(s_\tau,\tau,\ell),\Sigma_{\theta}(s_\tau,\tau,\ell)),
\end{align}
where $p_\theta(s_{\tau-1}|s_\tau,\ell)$ represents a parameterized Gaussian distribution, with its mean and variance provided by the trained noise predictor network $\mu_{\theta}(s_\tau,\tau,\ell), \Sigma_{\theta}(s_\tau,\tau,\ell)$. Therefore, how to design a noise predictor is the key to our generation performance, which we will discuss in detail in the following sections. The complete inference process is as follows:
\begin{align}
p_{\theta}(s_0|s_T,\ell) &= p(s_T)\prod^T_{\tau=1}p_{\theta}(s_{\tau-1}|s_\tau,\ell),
\end{align}
where $p(s_T)=\mathcal{N}(s_T;0,I)$, indicating that we start with isotropic Gaussian noise and progressively remove noise.

\subsubsection{Noise Predictor Network}
As previously mentioned, the key to the model's function lies in how to design a noise predictor for each phase. U-Net is an effective approach for learning spatial features from a given spatial map. It has an encoder with multiple parameterized down sampling layers to extract spatial features and a decoder with up sampling layers to project these features to the predicted noise.
\begin{gather}
\textbf{S} = {encoder}(\textbf{X}_{input}), \\
\hat{\epsilon} = {decoder}(\textbf{S}), \\
\hat{x_{\tau}} = x_\tau - \hat{\epsilon}.
\end{gather}
However, U-Net focuses on correlations within individual maps but overlooks the rich relationships across maps in adjacent history observations. We therefore aim to incorporate another network to handle the spatiotemporal information within the sequences. Specifically, we extract intra-map features from the current spatial map through the encoder of U-NET and inter-map features through another specialized encoder. These two features are then fused and fed into a decoder to predict the final noise. In each model, we have:
\begin{gather}
\textbf{S}_{intra} = {intra-encoder}(\textbf{X}_{\text{input}}^t), \\
\textbf{S}_{inter} = {inter-encoder}(\textbf{X}_{\text{input}}^{1:t}), \\
\hat{\epsilon} = {decoder}(\textbf{S}_{intra} \oplus \textbf{S}_{inter}).
\end{gather}

Moreover, since we utilize two diffusion models to handle coarse-grained completion and fine-grained inference, the noise predictor for the two stages should be customized for the task-specialized characteristics. We will introduce the encoder in two stages in the following subsections.

\begin{figure}[!t]
\centerline{\includegraphics[width=1.0\linewidth]{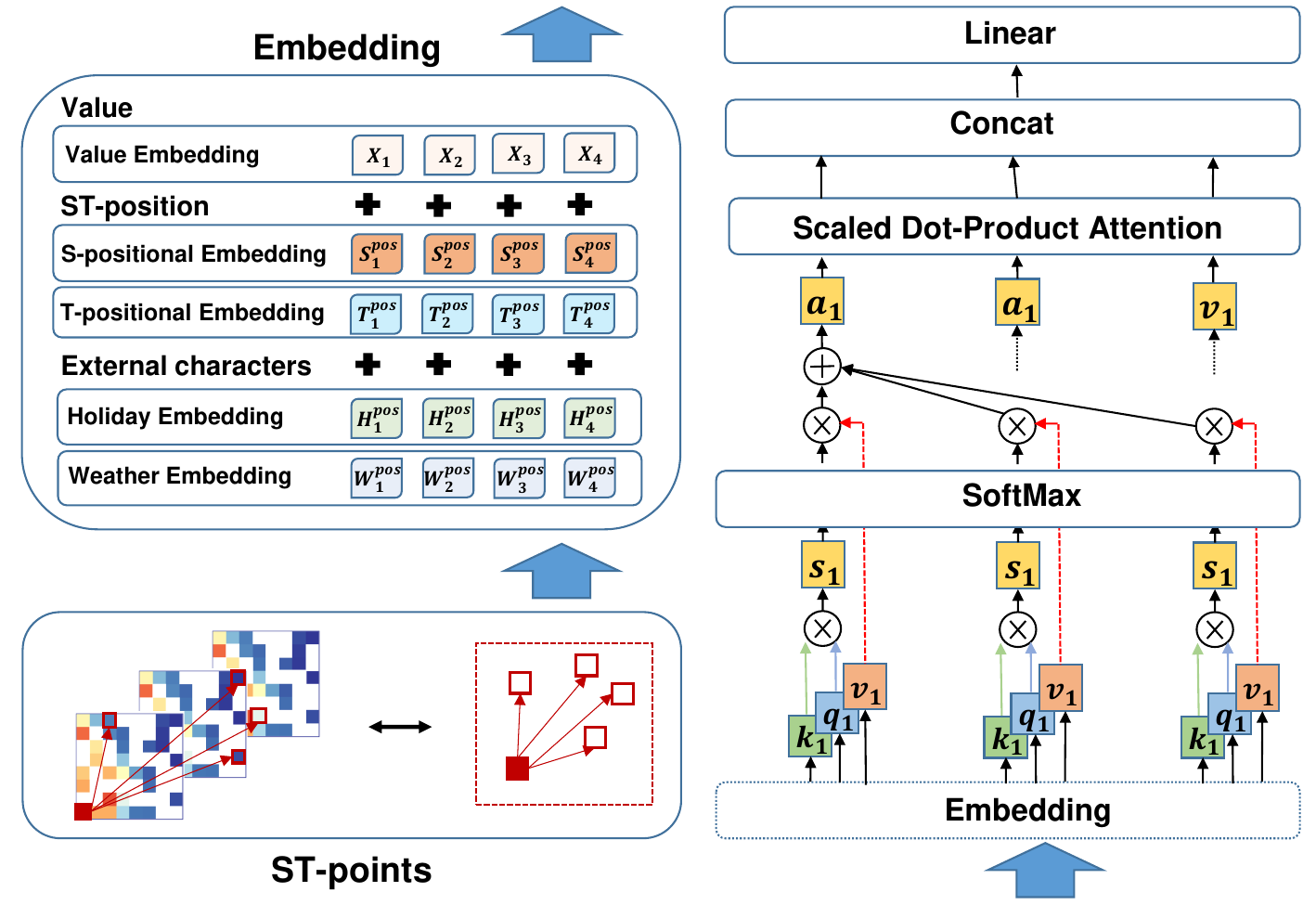}}
\caption{ The structure of ST-PointFormer. }
\label{fig:ST-PointFormer}
\vspace{-15pt}
\end{figure}

\subsection{ST-PointFormer}
\label{ST-PointFormer}
We observe that at this stage, the granularity of data remains consistent, and history data can provide direct help for completing missing data. For example, if there are recent observations at or near the locations where current data is missing, the historical observed values will be very close to the missing values we aim to infer. Therefore, we want to learn the complicated relationships between each spatiotemporal point (ST-point) in $\{\textbf{X}^i_{comp, cg}\}_{i = 0 \sim t}$ and utilize this information to help complete missing map $\textbf{X}^t_{comp, cg}$.

Inspired by \cite{wang2023spatiotemporal}, which brings attention mechanism as a powerful tool into spatiotemporal data processing, we propose the ST-Pointformer to encode history information by computing the relations between ST-points of history sequences. The structure of ST-PointFormer is shown in Figure~\ref{fig:ST-PointFormer}. We suppose that the relationships between ST-points are mainly decided by the observed value, the relative spatiotemporal position, and the external characteristics. Therefore, we first use three types of embedding layers to embed entering data and external knowledge into the model. 

We utilize the value embedding layer to project the data into $d_{model}$-dimension vector $\textbf{X}_{value} \in \mathbb{R}^{t \times I \times J \times d_{model}}$ through a fully connected layer. We also use a learnable position embedding layer to provide relative spatial and temporal relationships, denoted as $\textbf{X}_{pos} \in \mathbb{R}^{t \times I \times J \times d_{model}}$.  Another fully-connected layer is designed to learn from external features like daily, weekly, monthly indices and holiday information,which are embedded to $\textbf{X}_{external} \in \mathbb{R}^{t \times I \times J \times d_{model}}$. These embedding values will then be added together to enter the encoder.
\begin{align}
\textbf{X}_{feed, 0} &= \mathcal{E}(\textbf{X}) = \textbf{X}_{value} + \textbf{X}_{pos} + \textbf{X}_{external}.
\end{align}

After the observed data sequences are embedded to $\textbf{X}_{feed, 0}$, we use multi-head self-attention to compute the associations between ST-points, thus enabling further encoding for each ST-point. Self-attention is a particular implementation of attention mechanism where the query vector, key vector, and value vector are projected from the same data, and the multi-head attention helps the model capture richer information within the ST-point sequences. When the ST-point sequences $\textbf{X}_{embed}$ reach the attention layer, due to our multiple-head attention designed to capture spatiotemporal relationships of different possible patterns, for each attention head $head_i$, we use three learnable weight matrices $\textbf{W}_i^q$, $\textbf{W}_i^k$, and $\textbf{W}_i^v$ to project the embedded data to form three vectors, including the query vector $\textbf{Q}_i$,  key vector $\textbf{K}_i$, and  value vector $\textbf{V}_i$.
\begin{align}
\textbf{Q}_i = \textbf{X}_{embed} \textbf{W}_i^q,\textbf{K}_i = \textbf{X}_{embed} \textbf{W}_i^k,\textbf{V}_i = \textbf{X}_{embed} \textbf{W}_i^v.
\end{align}

After that, we use the formula of Scaled Dot-Product Attention to calculate the attention scores $\textbf{S}_i$.
\begin{align}
\textbf{S}_i &= {softmax}\left(\frac{\textbf{Q}_i \textbf{K}_i^T}{\sqrt{d_k}}\right).
\end{align}
The multiplication of $\textbf{Q}_i$ and $\textbf{K}_i$ indicates the similarity of them, i.e., the similarity of the spatiotemporal correlations between ST-points. The softmax function normalizes this similarity so that we make the attention scores between $0$ and $1$, where $\sqrt{d_k}$ prevents the gradient instability. Finally, we multiply the attention score matrix with $\textbf{V}_i$ to obtain the new ST-point representation which combines the information of all other ST-points considering their relationships.
\begin{align}
A(\textbf{Q}_i, \textbf{K}_i, \textbf{V}_i) &= \textbf{S}_i \textbf{V}_i.
\end{align}

Once the outputs of all attention heads have been computed, they are concatenated into a single matrix and projected through a learnable weight matrix $\textbf{W}_o$ to reduce its dimension.
\begin{align}
MA(\textbf{Q}, \textbf{K}, \textbf{V}) &= {Concat}(h_1, h_2, \dots, h_k) \textbf{W}_o, \\
h_i &= A(\textbf{Q}_i, \textbf{K}_i, \textbf{V}_i).
\end{align}
The encoder can have multiple attention layers inside. For each attention layer, it is passed through the self-attention layer to calculate the attention and perform residual connection and normalization, and then enters the feedforward network layer to improve the fitting ability by linear transformation.
\begin{align}
\textbf{X}_{feed, 0} &=  \mathcal{E}(\textbf{X}), \\ 
\textbf{X}_{attn, i} &= \mathcal{LN}(\mathcal{MA}(\textbf{X}_{feed, i}) + \mathcal{ADD}), \\
\textbf{X}_{feed, i+1} &= {ReLu}({ReLu}(\textbf{X}_{attn, i} \textbf{W}_{1,i}) \textbf{W}_{2,i}), \\
\textbf{X}_{out} &= {ReLu}({ReLu}(\textbf{X}_{attn, k} \textbf{W}_{1,k}) \textbf{W}_{2,k}),
\end{align}
where $\text{LN}(\cdot)$, $\text{MA}{(\cdot)}$, and $\text{ADD}$ represent the LayerNorm, the multi-attention layer, and residual connection. The $\text{RELU}$ is the activation function, and $\textbf{W}_1$, $\textbf{W}_2$ are the learnable weights.

Thus far, we have encoded each ST-point, allowing these ST-points to fully incorporate information from other relevant ST-points. The following step is to align these encoded values with the U-NET encoding results to perform the final inference. Typically, the output of U-NET $z_{spatial} \in \mathbb{R}^{I_{down} \times J_{down} \times d_{model}}$ is a downsampled spatial map with the length of $I_{down}$, width of $J_{down}$ and $d_{model}$ channels. At the same time, we have $I \times J \times t$ ST-points, each with a dimension of $d_{model}$. Since in $z_{intra}$ each position represents the extracted spatial semantic information of a subregion of size $\frac{I \times J}{I_{down} \times J_{down}}$ in the original spatial map, we stack encoded representation of ST-points within each subregion to maintain semantic consistency with the U-NET results. Specifically, for the top-left subregion, we have:
\begin{align}
\textbf{S}_{inter}^{[0,0]} &= {Concat}\left(\textbf{X}_{attn, K}^{[0,0]}, \cdots, \textbf{X}_{attn, K}^{[0,J_{down}-1]}, \right. \notag \\
&\quad \left. \textbf{X}_{attn, K}^{[1,0]}, \cdots, \textbf{X}_{attn, K}^{[1,J_{down}-1]}, \cdots, \right. \notag \\
&\quad \left. \textbf{X}_{attn, K}^{[I_{down}-1,0]}, \cdots, \textbf{X}_{attn, K}^{[I_{down}-1,J_{down}-1]}\right).
\end{align}
In this way, we obtain the final encoded results.
\begin{align}
\textbf{S}_{inter} &= {Stack}(\textbf{X}_{attn, K}).
\end{align}

It is worth noting that, unlike traditional convolutional methods, the attention mechanism does not rely on neighborhood properties to capture relationships between data values. Instead, it computes attention scores between any two spatiotemporal location in the data map, making our method more resilient to different sparse patterns in observed data. Whether missing data locations shift over time or are unevenly distributed spatially, the performance remains stable. In contrast, our experiments show that non-attention-based methods, which depend on local continuity, struggle to maintain this stability. This comparison will be discussed further in Section ~\ref{EXPERIMENTS}.

\begin{figure}[!t]
\centerline{\includegraphics[width=1.0\linewidth]{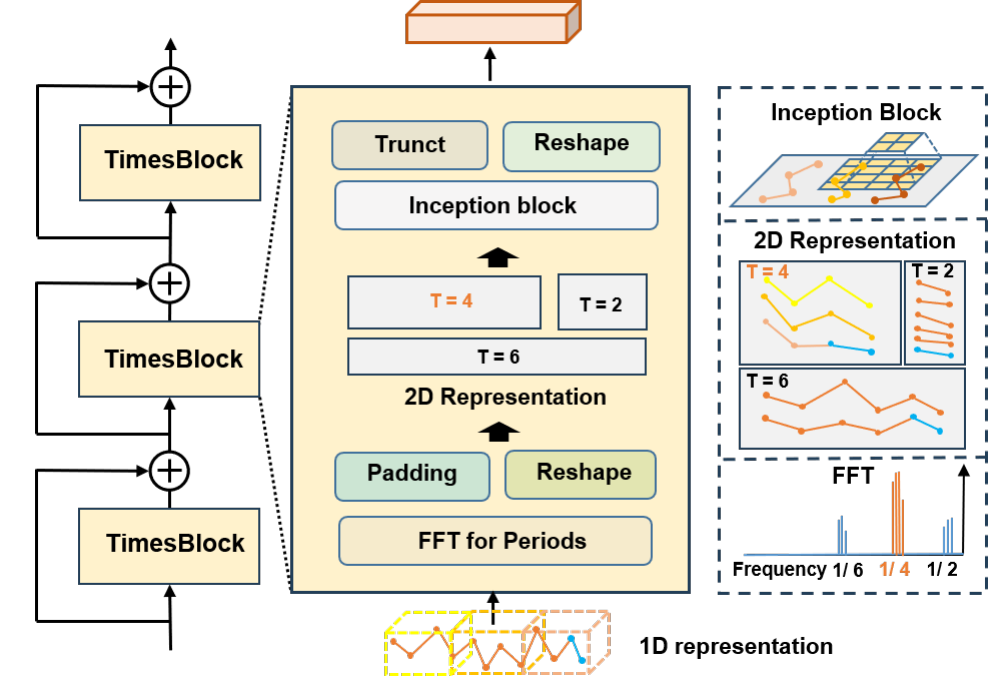}}
\caption{ The structure of T-PatternNet }
\label{fig:T-PatternNet}
\vspace{-15pt}
\end{figure}

\subsection{T-PatternNet}
\label{T-PatternNet}
Temporal patterns, such as periodicity and trends, are essential in spatiotemporal data processing and can significantly aid in data reconstruction. However, capturing these patterns in sparse is challenging, as missing values can disrupt temporal inferences. To address this, we defer the use of temporal patterns until the fine-grained inference stage, where the missing data has been approximated at a coarser level. At this point, our goal is to extract temporal patterns from the now-complete coarse-grained data and leverage them as key features to enhance fine-grained inference. Inspired by \cite{wu2022timesnet}, we utilize T-PatternNet to analyze the completed results of history data $\{\textbf{X}^i_{comp, cg}\}_{i = 0 \sim t}$ as a 1-D sequential series with $I \times J$ channels and a length of $t$.

As illustrated in Figure~\ref{fig:T-PatternNet}, T-PatternNet consists of multiple TimesBlocks that analyze two types of temporal variations: variations between adjacent areas and those with the same phase across different periods, referred to as intraperiod- and interperiod- variations. To facilitate this, the 1-D temporal series is transformed into 2-D data after identifying its intrinsic periodicity using Fast Fourier Transformer (FFT). Following this, an inception block with multi-scale 2-D kernels aggregates the intraperiod variations (columns) and interperiod variations (rows) simultaneously. The aggregated representation is then fused, with their importance determined by the FFT results.

Specifically, with the complete coarse-grained results from the coarse-grained completion stage, the input to T-PatternNet is $\textbf{X} \in \mathbb{R}^{t \times (I\times J)}$. We first embed $\textbf{X}$ into deep features using a linear embedding layer:
\begin{equation}
    \textbf{X}_{1D} = \mathcal{E}\textbf{(X}) = \textbf{X} \times \textbf{W}^{1D},
\end{equation}
where $\textbf{X}_{1D} \in \mathbb{R}^{t \times d_{model}}$ and $\textbf{W}_{1D} \in \mathbb{R}^{(I \times J) \times d_{model}}$. Then we transform the embedded data into the frequency domain by FFT to analyze its inner periodicity.
\begin{align}
    \textbf{A} &= {Avg}({Amp}({FFT}(\textbf{X}_{1D}))), \\
    \{f_1, \dots, f_k\} &= \mathop{\arg {Topk}}\limits_{f_* \in \{1, \dots, [\frac{T}{2}]\}}(\textbf{A}), \\
    p_i &= \left\lceil \frac{T}{f_i} \right\rceil, \quad i \in \{1, \dots, k\}.
\end{align}
Here, $\text{FFT}(\cdot)$ and $\text{Amp}(\cdot)$ denote the FFT and the calculation of amplitude values. $\textbf{A} \in \mathbb{R}^t$ represents the averaged amplitude of each frequency. As \cite{wu2022timesnet}, we only select top-$k$ amplitudes to avoid noise, where $k$ is the hyper-parameter, and the selected frequencies correspond to the preferred periodicity. For each selected periodicity $p_i$, we fold the 1-D series $\textbf{X}^{1D}$ to form the 2-D tensors with $p_i$ rows and $f_i$ columns.
\begin{equation}
    \textbf{X}_{2D} = {Reshape}_{p_i, f_i}({Padding}(\textbf{X}_{1D})).
\end{equation}
Here, the $\text{Padding}(\cdot)$ represents extending series by zeros to make it compatible for $\text{Reshape}_{p_i, f_i}$.

After we obtain the 2-D tensors, the next step is to extract the intraperiod- and interperiod-variations. This is done by an inception block, namely $\text{Inception}(\cdot)$, which contains multi-scale 2-D kernels and is pervasively used in computer visions as a powerful feature extractor. After the 2-D tensors pass through the inception block, we use another $\text{Trunc}(\cdot)$ to truncate the padded series into the original length $n$.
\begin{align}
    &\hat{\textbf{X}_{2D}^{l,i}} = {Inception}(\textbf{X}_{2D}^{l,i}), \quad i \in \{1, \dots, k\}, \\
    &\hat{\textbf{X}_{1D}^{l,i}} = {Trunc}({Reshape}_{1, (p_i \times f_i)}(\hat{\textbf{X}_{2D}^{l,i}})), i \in \{1, \dots, k\}.
\end{align}

Finally, we fuse the $k$ different 1-D representations according to their relative importance, which is decided by the amplitudes calculated at the FFT stage. The fused representation from the $(l-1)-th$TimesBlock  will be sent to the next $l-th$ TimesBlock for deeper feature extraction.
\begin{align}
    \hat{A_{f_1}^{l-1}}, \dots, \hat{A_{f_k}^{l-1}} &= {softmax}(A_{f_1}^{l-1}, \dots, A_{f_k}^{l-1}), \\
    \textbf{X}_{1D}^l &= \sum\limits_{i=1}^{k} \hat{A_{f_i}^{l-1}} \times \hat{\textbf{X}_{1D}^{l,i}}.
\end{align}
As in the stacking process in ST-PointFormer, we obtain the final encoded results by stacking the $\textbf{X}_{1D}^{L}$ representation of all the spatial locations.

\subsection{Training Strategy}
\label{Training Strategy}
In DiffRecon, the input consists of sparse coarse-grained data, while the ground truth is the complete fine-grained data. We begin by pre-training each stage of the model separately to ensure that each component is optimized for its specific role in the reconstruction process. During this pre-training phase, we downsample the fine-grained data to create a pseudo-complete coarse-grained version, which serves as the ground truth for Diffusion-C and is used as the input for Diffusion-F. This strategy allows both stages to learn how to handle the incomplete data independently. Once pre-training is complete, we proceed with joint training, where both stages are integrated into a single end-to-end framework. This joint training phase refines the model’s ability to seamlessly connect the stages and perform more accurate reconstructions of fine-grained data from sparse coarse-grained inputs.

\begin{table*}[htbp]
  \centering
  \caption{Dataset Description}
  \setlength{\tabcolsep}{15pt} 
    \begin{tabular}{lccc}
    \toprule
    \textbf{Dataset} & \textbf{TaxiBJ} & \textbf{BikeNYC} & \textbf{TAPBJ} \\
    \midrule
    Latitude & (39.8200,39.9966) & (40.71,40.765) & (39.4,41.1) \\
    Longitude & (116.2498,116.4950) & (-74.01,-73.972) & (115.4,117.5) \\
    Time span & Jul 1, 2013 - Oct 30, 2013 & Jan 1, 2016 - Apr 30, 2016 & Jan 1, 2014 - Dec 31, 2017 \\
    Time interval & 30 minutes & 1 h   & 1 day \\
    Coarse/Fine-grained shape & 8×8/32×32 & 8×8/32×32 & 8×8/32×32 \\
    External features & \checkmark    & \checkmark    & \checkmark \\
    \bottomrule
    \end{tabular}%
  \label{tab:datasets}%
\end{table*}%

\section{EXPERIMENTS}
\label{EXPERIMENTS}
\subsection{Experiment Setting}
\subsubsection{Dataset}
We use three datasets to validate the effectiveness of the proposed model: TaxiBJ, BikeNYC, and TAPBJ. The statistics of these datasets are provided in Table~\ref{tab:datasets}. Here is the detailed information about the datasets:

\begin{itemize}
    \item \textbf{TaxiBJ\cite{zhang2017deep}}: This dataset contains taxi traffic data in Beijing. We use the data collected every half hour from 7/1/2013 to 12/31/2013. The entire dataset is divided into non-overlapping training, validation, and test sets in a ratio of 2:1:1.
    
    \item \textbf{BikeNYC\footnote{https://ride.citibikenyc.com/system-data}}: This dataset includes millions of bike trip records in New York from 1/1/2016 to 4/30/2016. Each record contains various information such as Start Time, End Time, start station latitude/longitude, end station latitude/longitude, etc. We preprocess all data with a time interval of 1 hour and divide the dataset in a ratio of 2:1:1 for training, validation, and testing.
    
    \item \textbf{TAPBJ\footnote{http://tapdata.org.cn}\cite{geng2021tracking,xiao2022spatiotemporal,xiao2021separating}}: This dataset includes PM2.5 data from China since 2000, with a daily cycle. We use the data from Beijing from 1/1/2014 to 12/31/2017. Similar to the other datasets, we divide it into training, validation, and test sets in a ratio of 2:1:1.
    
    \item \textbf{External features}: The external features used include weather, temperature, humidity, holidays, etc.
\end{itemize}

\subsubsection{Data Preprocessing}
To test our model's performance in various real-world scenarios with missing data, we considered three task scenarios of typical sparse patterns:

\begin{itemize}
    \item In traditional scenarios using fixed sensors, the positions of observed and missing data are fixed. We considered missing ratios of 0.2, 0.4, 0.6, and 0.8. By proportionally masking values at fixed positions in each time slice, we fully tested our model's performance across different sparsity levels (from sparse to relatively complete).
    \item By recruiting users to collect data using mobile sensors, the positions of observed and missing data are random in each time slice. Similarly, we considered missing ratios of 0.2, 0.4, 0.6, and 0.8 and proportionally masked values at random positions in each time slice.
    \item For large-area data missing due to emergencies (e.g., power outages, natural disasters), we simulated these scenarios by masking the values in the lower right quarter of the coarse-grained data. Apart from the complete absence of data in the lower right quarter, we assume the data in other areas is intact.
\end{itemize}

\subsubsection{Evaluation Metrics}
We use Mean Absolute Error (MAE) and Root Mean Square Error (RMSE) to evaluate the performance of our model. The formulas are as follows: 
\begin{align}
\text{MAE} &= \frac{1}{T} \sum_{t=0}^{T} \left| \mathcal{X}^t - \mathcal{Y}^t \right|, \\
\text{RMSE} &= \sqrt{\frac{1}{T} \sum_{t=0}^{T} \left\| \mathcal{X}^t - \mathcal{Y}^t \right\|_F^2},
\end{align}
where $\mathcal{X}^t$ denotes the fine-grained spatiotemporal data distribution inferred by DiffRecon, and $\mathcal{Y}^t$ denotes the ground truth. Smaller metric scores indicate better model performance.

\begin{table*}[t!]
  \centering
  \caption{The performance results of DiffRecon and other baseline methods on the TaxiBJ, BikeNYC, and TAPBJ datasets in fixed and random data missing scenarios.}
    \renewcommand{\arraystretch}{1.2} 
    \begin{tabular}{cc|cc|c|cccccccccc}
    \specialrule{.15em}{.05em}{.05em} %
    \multicolumn{2}{c}{\multirow{2}[2]{*}{\textbf{Dataset}}} &       & \multicolumn{1}{r}{} & \multicolumn{1}{r}{} &       & \multirow{2}[2]{*}{\textbf{SRCNN}} & \multirow{2}[2]{*}{\textbf{SRResNet}} & \multirow{2}[2]{*}{\textbf{UrbanFM}} & \multirow{2}[2]{*}{\textbf{FODE}} & \multirow{2}[2]{*}{\textbf{STCF}} & \multirow{2}[2]{*}{\textbf{UrbanSTA}} & \multirow{2}[2]{*}{\textbf{DiffUFlow}} & \multirow{2}[2]{*}{\textbf{DiffRecon}} & \multirow{2}[2]{*}{\textbf{Improve}} \\
    \multicolumn{2}{c}{} &       & \multicolumn{1}{r}{} & \multicolumn{1}{r}{} &       &       &       &       &       &       &       &       &       &  \\
    \specialrule{.15em}{.05em}{.05em} %
    
    \multicolumn{2}{c|}{\multirow{16}[16]{*}{TaxiBJ}} & \multicolumn{2}{c|}{\multirow{8}[8]{*}{fix}} & \multirow{2}[2]{*}{20\%} & MAE   & 32.92  & 33.90  & 28.19  & 35.71  & 32.73  & \underline{27.52}  & 27.89  & \textbf{23.42}  & 14.9\%  \\
    \multicolumn{2}{c|}{} & \multicolumn{2}{c|}{} &       & RMSE  & 52.40  & 52.25  & 48.97  & 52.46  & 58.43  & \underline{42.24}  & 42.64  & \textbf{36.71}  & 13.1\%  \\
\cline{5-5}    \multicolumn{2}{c|}{} & \multicolumn{2}{c|}{} & \multirow{2}[2]{*}{40\%} & MAE   & 44.81  & 43.72  & 38.97  & 44.78  & 48.57  & \underline{30.40}  & 31.13  & \textbf{24.72}  & 18.7\%  \\
    \multicolumn{2}{c|}{} & \multicolumn{2}{c|}{} &       & RMSE  & 69.64  & 62.80  & 65.31  & 68.48  & 77.96  & 48.16  & \underline{47.94} & \textbf{39.70}  & 17.2\%  \\
\cline{5-5}    \multicolumn{2}{c|}{} & \multicolumn{2}{c|}{} & \multirow{2}[2]{*}{60\%} & MAE   & 52.54  & 54.20  & 53.54  & 62.53  & 66.12  & \underline{31.52}  & 33.98  & \textbf{26.12}  & 17.1\%  \\
    \multicolumn{2}{c|}{} & \multicolumn{2}{c|}{} &       & RMSE  & 77.97  & 83.58  & 82.71  & 86.90  & 102.56  & \underline{52.80}  & 54.39  & \textbf{42.49}  & 19.5\%  \\
\cline{5-5}    \multicolumn{2}{c|}{} & \multicolumn{2}{c|}{} & \multirow{2}[2]{*}{80\%} & MAE   & 68.81  & 69.84  & 76.97  & 73.89  & 84.00  & \underline{34.40}  & 38.70  & \textbf{27.83}  & 19.1\%  \\
    \multicolumn{2}{c|}{} & \multicolumn{2}{c|}{} &       & RMSE  & 94.96  & 95.66  & 108.46  & 106.35  & 118.01  & \underline{58.08}  & 62.87  & \textbf{45.36}  & 21.9\%  \\
\cline{3-15}    \multicolumn{2}{c|}{} & \multicolumn{2}{c|}{\multirow{8}[8]{*}{random}} & \multirow{2}[2]{*}{20\%} & MAE   & 35.68  & 38.08  & 33.76  & 42.08  & 29.83  & 60.00  & \underline{29.61}  & \textbf{24.22}  & 18.2\%  \\
    \multicolumn{2}{c|}{} & \multicolumn{2}{c|}{} &       & RMSE  & 58.56  & 60.16  & 58.40  & 64.96  & 52.00  & 98.24  & \underline{45.12}  & \textbf{38.39}  & 14.9\%  \\
\cline{5-5}    \multicolumn{2}{c|}{} & \multicolumn{2}{c|}{} & \multirow{2}[2]{*}{40\%} & MAE   & 48.32  & 51.04  & 47.04  & 54.72  & 42.70  & 91.52  & \underline{32.93}  & \textbf{27.08}  & 17.8\%  \\
    \multicolumn{2}{c|}{} & \multicolumn{2}{c|}{} &       & RMSE  & 75.36  & 77.44  & 74.24  & 82.56  & 70.00  & 129.92  & \underline{51.43}  & \textbf{41.66}  &19.0\%  \\
\cline{5-5}    \multicolumn{2}{c|}{} & \multicolumn{2}{c|}{} & \multirow{2}[2]{*}{60\%} & MAE   & 64.07  & 65.76  & 62.08  & 69.44  & 58.89  & 120.80  & \underline{34.17}  & \textbf{30.06}  & 12.0\%  \\
    \multicolumn{2}{c|}{} & \multicolumn{2}{c|}{} &       & RMSE  & 93.30  & 95.04  & 91.36  & 100.16  & 88.77  & 156.48  & \underline{54.60}  & \textbf{49.81}  & 8.8\%  \\
\cline{5-5}    \multicolumn{2}{c|}{} & \multicolumn{2}{c|}{} & \multirow{2}[2]{*}{80\%} & MAE   & 88.48  & 91.04  & 88.00  & 92.64  & 84.70  & 138.08  & \underline{38.55}  & \textbf{35.97}  & 6.7\%  \\
    \multicolumn{2}{c|}{} & \multicolumn{2}{c|}{} &       & RMSE  & 120.48  & 123.68  & 121.28  & 127.52  & 118.20  & 172.16  & \underline{61.64}  & \textbf{60.48}  & 1.9\%  \\
    
    \specialrule{.15em}{.05em}{.05em} %
    \multicolumn{2}{c|}{\multirow{16}[16]{*}{BikeNYC}} & \multicolumn{2}{c|}{\multirow{8}[8]{*}{fix}} & \multirow{2}[2]{*}{20\%} & MAE   & 0.47  & 0.44  & 0.34  & 0.32  & \textbf{0.28}  & 0.32  & \underline{0.29}  & 0.30  & -7.1\%  \\
    \multicolumn{2}{c|}{} & \multicolumn{2}{c|}{} &       & RMSE  & 1.08  & 1.04  & 1.23  & 1.09  & 1.10  & 1.17  & \textbf{0.89}  & \underline{0.90}
  & -1.1\%  \\
\cline{5-5}    \multicolumn{2}{c|}{} & \multicolumn{2}{c|}{} & \multirow{2}[2]{*}{40\%} & MAE   & 0.48  & 0.48  & 0.44  & 0.36  & 0.38  & 0.36  & \textbf{0.32}  & \underline{0.34}  & -6.3\%  \\
    \multicolumn{2}{c|}{} & \multicolumn{2}{c|}{} &       & RMSE  & 1.12  & 1.11  & 1.48  & 1.18  & 1.44  & 1.23  & \underline{1.02}  & \textbf{0.97}  & 4.9\%  \\
\cline{5-5}    \multicolumn{2}{c|}{} & \multicolumn{2}{c|}{} & \multirow{2}[2]{*}{60\%} & MAE   & 0.62  & 0.54  & 0.43  & 0.43  & 0.44  & 0.40  & \underline{0.33}  & \textbf{0.28}  & 15.1\%  \\
    \multicolumn{2}{c|}{} & \multicolumn{2}{c|}{} &       & RMSE  & 1.33  & 1.19  & 1.42  & 1.49  & 1.67  & 1.47  & \underline{1.07}  & \textbf{1.05}  & 1.9\%  \\
\cline{5-5}    \multicolumn{2}{c|}{} & \multicolumn{2}{c|}{} & \multirow{2}[2]{*}{80\%} & MAE   & 0.74  & 0.62  & 0.50  & 0.48  & 0.48  & 0.42  & \underline{0.38}  & \textbf{0.36}  & 5.3\%  \\
    \multicolumn{2}{c|}{} & \multicolumn{2}{c|}{} &       & RMSE  & 1.57  & 1.39  & 1.81  & 1.68  & 1.68  & 1.55  & \textbf{1.16}  & \underline{1.20}  & -3.4\%  \\
\cline{3-15}    \multicolumn{2}{c|}{} & \multicolumn{2}{c|}{\multirow{8}[8]{*}{random}} & \multirow{2}[2]{*}{20\%} & MAE   & 0.47  & 0.55  & 0.37  & 0.32  & \underline{\textbf{0.28}}  & 0.47  & 0.30  & \underline{\textbf{0.28}}  & 0\%  \\
    \multicolumn{2}{c|}{} & \multicolumn{2}{c|}{} &       & RMSE  & 1.18  & 1.27  & 1.32  & 1.14  & 1.04  & 1.32  & \textbf{0.91}  & \underline{0.94}  & -3.3\%  \\
\cline{5-5}    \multicolumn{2}{c|}{} & \multicolumn{2}{c|}{} & \multirow{2}[2]{*}{40\%} & MAE   & 0.52  & 0.51  & 0.39  & 0.38  & \underline{0.34}  & 0.63  & 0.35  & \textbf{0.27}  & 20.6\%  \\
    \multicolumn{2}{c|}{} & \multicolumn{2}{c|}{} &       & RMSE  & 1.31  & 1.27  & 1.36  & 1.33  & 1.22  & 1.52  & \underline{1.05}  & \textbf{1.00}  & 4.8\%  \\
\cline{5-5}    \multicolumn{2}{c|}{} & \multicolumn{2}{c|}{} & \multirow{2}[2]{*}{60\%} & MAE   & 0.61  & 0.68  & 0.45  & 0.45  & 0.41  & 0.75  & \underline{0.37}  & \textbf{0.31}  & 16.2\%  \\
    \multicolumn{2}{c|}{} & \multicolumn{2}{c|}{} &       & RMSE  & 1.46  & 1.71  & 1.55  & 1.52  & 1.41  & 1.65  & \underline{1.12}  & \textbf{1.11}  & 0.9\%  \\
\cline{5-5}    \multicolumn{2}{c|}{} & \multicolumn{2}{c|}{} & \multirow{2}[2]{*}{80\%} & MAE   & 0.79  & 0.69  & 0.52  & 0.52  & 0.49  & 0.79  & \underline{0.46}  & \textbf{0.33} & 28.3\%  \\
    \multicolumn{2}{c|}{} & \multicolumn{2}{c|}{} &       & RMSE  & 1.83  & 2.05  & 1.75  & 1.73  & 1.63  & 1.80  & \underline{1.59}  & \textbf{1.23}  & 22.6\%  \\
    \specialrule{.15em}{.05em}{.05em} %
    
    \multicolumn{2}{c|}{\multirow{16}[16]{*}{TAPBJ}} & \multicolumn{2}{c|}{\multirow{8}[8]{*}{fix}} & \multirow{2}[2]{*}{20\%} & MAE   & 2.77  & 2.86  & \underline{2.57}  & 3.27  & 2.87  & 3.04  & 2.74  & \textbf{1.77}  & 31.1\%  \\
    \multicolumn{2}{c|}{} & \multicolumn{2}{c|}{} &       & RMSE  & 4.69  & 4.91  & 4.44  & 5.55  & 5.73  & 4.93  & \underline{4.33}  & \textbf{3.00}  & 30.7\%  \\
\cline{5-5}    \multicolumn{2}{c|}{} & \multicolumn{2}{c|}{} & \multirow{2}[2]{*}{40\%} & MAE   & 3.93  & 3.65  & 4.02  & 4.24  & 3.67  & 4.50  & \underline{3.41}  & \textbf{2.14}  & 37.2\%  \\
    \multicolumn{2}{c|}{} & \multicolumn{2}{c|}{} &       & RMSE  & 7.22  & 6.53  & 8.59  & 7.63  & 6.91  & 6.57  & \underline{5.32}  & \textbf{3.52}  & 33.8\%  \\
\cline{5-5}    \multicolumn{2}{c|}{} & \multicolumn{2}{c|}{} & \multirow{2}[2]{*}{60\%} & MAE   & 5.20  & 4.89  & 5.32  & 5.48  & 4.98  & 9.27  & \underline{4.64}  & \textbf{3.02}  & 34.9\%  \\
    \multicolumn{2}{c|}{} & \multicolumn{2}{c|}{} &       & RMSE  & 9.44  & 8.65  & 10.13  & 10.01  & 9.13  & 14.92  & \underline{6.99}  & \textbf{5.10}  & 27.0\%  \\
\cline{5-5}    \multicolumn{2}{c|}{} & \multicolumn{2}{c|}{} & \multirow{2}[2]{*}{80\%} & MAE   & 7.87  & 8.92  & 8.82  & 9.03  & 9.86  & 9.63  & \underline{6.11}  & \textbf{4.29} & 29.8\%  \\
    \multicolumn{2}{c|}{} & \multicolumn{2}{c|}{} &       & RMSE  & 13.91  & 16.26  & 15.84  & 16.27  & 18.94  & 15.19  & \underline{9.19}  & \textbf{6.77}  & 26.3\%  \\
\cline{3-15}    \multicolumn{2}{c|}{} & \multicolumn{2}{c|}{\multirow{8}[8]{*}{random}} & \multirow{2}[2]{*}{20\%} & MAE   & 2.83  & 2.89  & 2.69  & 3.19  & \underline{2.33}  & 3.98  & 2.98  & \textbf{1.99}  & 14.6\%  \\
    \multicolumn{2}{c|}{} & \multicolumn{2}{c|}{} &       & RMSE  & 4.82  & 4.90  & 4.64  & 5.35  & \underline{4.29}  & 7.69  & 4.75  & \textbf{3.43}  & 20.0\%  \\
\cline{5-5}    \multicolumn{2}{c|}{} & \multicolumn{2}{c|}{} & \multirow{2}[2]{*}{40\%} & MAE   & 5.32  & 3.66  & 3.51  & 3.93  & \underline{3.34}  & 4.74  & 3.71  & \textbf{2.68}  & 19.8\%  \\
    \multicolumn{2}{c|}{} & \multicolumn{2}{c|}{} &       & RMSE  & 8.31  & 6.43  & 6.21  & 6.80  & 6.16  & 9.05  & \underline{5.88}  & \textbf{4.60}  & 21.8\%  \\
\cline{5-5}    \multicolumn{2}{c|}{} & \multicolumn{2}{c|}{} & \multirow{2}[2]{*}{60\%} & MAE   & 5.22  & 5.14  & 5.05  & 5.37  & 4.96  & 5.47  & \underline{4.72}  & \textbf{3.90}  & 17.4\%  \\
    \multicolumn{2}{c|}{} & \multicolumn{2}{c|}{} &       & RMSE  & 9.36  & 9.35  & 9.40  & 9.48  & 9.28  & 10.92  & \textbf{7.23}  & \underline{8.04}  & -11.2\%  \\
\cline{5-5}    \multicolumn{2}{c|}{} & \multicolumn{2}{c|}{} & \multirow{2}[2]{*}{80\%} & MAE   & 8.70  & 8.84  & 8.78  & 9.09  & 8.97  & 9.28  & \underline{6.20}  & \textbf{5.97}  & 3.7\%  \\
    \multicolumn{2}{c|}{} & \multicolumn{2}{c|}{} &       & RMSE  & 15.71  & 16.24  & 16.00  & 16.79  & 16.60  & 18.25  & \textbf{9.23}  & \underline{9.46}  & -2.5\%  \\
    \specialrule{.15em}{.05em}{.05em} %
    \end{tabular}%
  \label{tab:DiffRecon_result}%
\end{table*}%

\begin{table*}[t]
  \centering
  \caption{The performance results of DiffRecon and other baseline methods on the TaxiBJ, BikeNYC, and TAPBJ datasets in large-scale(LS) data missing scenarios.}
    \renewcommand{\arraystretch}{1.2} 
    \begin{tabular}{cc|cc|c|lccccccccc}
    \specialrule{.15em}{.05em}{.05em} %
    \multicolumn{2}{c}{\multirow{2}[2]{*}{\textbf{Dataset}}} &       & \multicolumn{1}{r}{} & \multicolumn{1}{r}{} &       & \multirow{2}[2]{*}{\textbf{SRCNN}} & \multirow{2}[2]{*}{\textbf{SRResNet}} & \multirow{2}[2]{*}{\textbf{UrbanFM}} & \multirow{2}[2]{*}{\textbf{FODE}} & \multirow{2}[2]{*}{\textbf{STCF}} & \multirow{2}[2]{*}{\textbf{UrbanSTA}} & \multirow{2}[2]{*}{\textbf{DiffUFlow}} & \multirow{2}[2]{*}{\textbf{DiffRecon}} & \multirow{2}[2]{*}{\textbf{Improve}} \\
    \multicolumn{2}{c}{} &       & \multicolumn{1}{r}{} & \multicolumn{1}{r}{} &       &       &       &       &       &       &       &       &       &  \\
    \specialrule{.15em}{.05em}{.05em} %
    \multicolumn{2}{c|}{\multirow{2}[2]{*}{TaxiBJ}} & \multicolumn{2}{c|}{\multirow{6}[6]{*}{ LS }} & \multirow{6}[6]{*}{25\%} & MAE   & 57.38  & 58.81  & 53.96  & 61.58  & 52.32  & \underline{28.96}  & 31.64  & \textbf{23.98}  & 17.2\%  \\
    \multicolumn{2}{c|}{} & \multicolumn{2}{c|}{} &       & RMSE  & 97.79  & 98.66  & 95.83  & 100.30  & 95.30  & \underline{47.20}  & 48.45  & \textbf{37.54}  & 20.5\%  \\
    \cline{1-2} \cline{6-15}
    \multicolumn{2}{c|}{\multirow{2}[2]{*}{BikeNYC}} & \multicolumn{2}{c|}{} &       & MAE   & 0.50  & 0.56  & 0.35  & 0.35  & 0.32  & \underline{0.29}  & 0.32  & \textbf{0.27}  & 6.9\%  \\
    \multicolumn{2}{c|}{} & \multicolumn{2}{c|}{} &       & RMSE  & 1.22  & 1.28  & 1.30  & 1.27  & 1.19  & 1.12  & \textbf{0.96}  & \underline{0.98}  & -2.1\%  \\
    \cline{1-2} \cline{6-15}
    \multicolumn{2}{c|}{\multirow{2}[2]{*}{TAPBJ}} & \multicolumn{2}{c|}{} &       & MAE   & 5.58  & 5.64  & 5.50  & 5.91  & 5.15  & 4.97  & \underline{4.27}  & \textbf{2.96}  & 30.7\%  \\
    \multicolumn{2}{c|}{} & \multicolumn{2}{c|}{} &       & RMSE  & 13.12  & 13.17  & 13.09  & 13.30  & 12.97  & 11.21  & \underline{8.53}  & \textbf{6.00}  & 29.7\%  \\
    \specialrule{.15em}{.05em}{.05em} %
    \end{tabular}%
  \label{tab:LargeScale-Result}%
\end{table*}%

\subsubsection{Baselines}

We compare DiffRecon with seven baseline methods, including traditional super-resolution reconstruction methods SRCNN and SRResNet, which are mainly applied to image super-resolution reconstruction without considering the characteristics of spatiotemporal data; classic fine-grained spatiotemporal data (especially urban flow data) inference methods UrbanFM, FODE, and STCF, which consider the characteristics of spatiotemporal data, such as external features, and aim to obtain fine-grained spatiotemporal data from coarse-grained spatiotemporal data. These methods do not consider data sparsity, and some works directly set missing values to -1, while we use a simple nearest neighbor interpolation algorithm for preprocessing. The latest fine-grained spatiotemporal data inference methods that consider data missing problems are UrbanSTA and DiffUFlow, which achieve SOTA in the task of inferring from incomplete coarse-grained data to complete fine-grained data. However, they are mainly used for scenarios with fixed missing positions. The specific baselines are introduced below:

\begin{itemize}
    \item \textbf{SRCNN \cite{dong2015image}}: SRCNN, as an end-to-end algorithm, first applied convolutional neural networks to the image super-resolution reconstruction task. It consists of three layers: patch extraction, non-linear mapping, and reconstruction.
    \item \textbf{SRResNet \cite{ledig2017photo}}: SRResNet uses a residual architecture to stack deeper network layers for image super-resolution.
    \item \textbf{UrbanFM \cite{liang2019urbanfm}}: The first work to study the Fine-grained Urban Flow Inference (FUFI) problem, containing three key modules: an inference network for learning spatial correlations, a distributed upsampling layer for applying spatial constraints, and an external feature fusion subnet.
    \item \textbf{FODE \cite{zhou2020enhancing}}: The authors present a new method with Ordinary Differential Equations (ODEs) to alleviate the large parameter updates and memory cost of CNNs, while considering the impact of external factors.
    \item \textbf{STCF \cite{xu2023spatial}}: A Spatiotemporal Contrasting for Fine-grained urban flow inference (STCF) method, which mitigates the overfitting problem by performing spatiaotemporal contrasting and feature fusion through two pre-trained networks and a fine-tuning network. Currently the SOTA method for the FUFI problem.
    \item \textbf{UrbanSTA \cite{wang2023urbanSTA}}: Considering the data missing problem in spatiotemporal fine-grained inference, it assumes that the missing data positions are fixed and proposes a multi-task framework named UrbanSTA, which designs a completion network (STA) and a fine-grained inference network (FIN) to fully consider spatiotemporal features through a spatiotemporal attention encoder.
    \item \textbf{DiffUFlow \cite{zheng2023diffuflow}}: Aiming at the problems of data missing and unreliable observation data, it proposes STFormer and ELFetcher to extract spatiotemporal features and semantic features respectively within the Diffusion model framework to achieve fine-grained inference. It is currently the SOTA method for inferring from incomplete coarse-grained to complete fine-grained data.
    
\end{itemize}

\subsection{Comparison With Baselines}
    
The performance results of DiffRecon and other baseline methods on the TaxiBJ, BikeNYC, and TAPBJ datasets are presented in Table~\ref{tab:DiffRecon_result},~\ref{tab:LargeScale-Result}. The percentages 20\%, 40\%, 60\%, and 80\% represent the proportion of missing coarse-grained data. "Fix" indicates that the missing data locations are fixed across all time slices, while "Random" indicates that the missing data locations are random for each time slice. The best performance in each row is highlighted in bold, and the second-best is underlined.

In scenarios with fixed data collection (fixed masks), SRCNN and SRResNet perform worse because they are designed for image super-resolution problems. They consider image deblurring issues but neglect the inherent characteristics of spatiotemporal data. UrbanFM, FODE, and STCF, although considering the spatiotemporal correlations of the data, ignore the impact of data sparsity, resulting in poor performance. UrbanSTA and DiffUFlow, as SOTA methods for inferring complete fine-grained data from incomplete coarse-grained data, perform significantly better than the above methods. In the first stage of DiffRecon, the spatiotemporal relationships between the collected incomplete coarse-grained spatiotemporal data are thoroughly considered. In the second stage, the periodicity and trend of the spatiotemporal data are further examined based on the complete coarse-grained data obtained from the first stage. By combining the first and second stages through pre-training and joint training strategies, the model generally performs the best in most cases. 

In scenarios with random data collection (random masks), the performance of baseline methods generally declines compared to fixed masks. This is because when the positions of collected data are random, spatiotemporal relationships become complex and difficult to exploit. Furthermore, when the data missing pattern is consistent for each time slice, the model fitting is easier; however, when the data missing pattern is random for each time slice, the model fitting becomes much harder. Taking the BikeNYC dataset as an example, with 80\% of the data missing, when using a fixed mask, DiffUFlow has an RMSE of 1.16, whereas with a random mask, the RMSE rises to 1.59. Notably, the performance of UrbanSTA declines the most significantly, often even falling short compared to methods that do not address the issue of data sparsity. This is because it assumes fixed missing data positions when utilizing spatiotemporal relationships, and its carefully designed spatiotemporal attention learning mechanism fails, leading to poor performance in random missing data scenarios.

In scenarios with large-scale missing data (large-scale mask), the performance of most baseline methods significantly declines compared to fixed data collection scenarios. For example, the MAE and RMSE of UrbanFM are 38.97 and 65.31, respectively, under the fixed mask with 40\% data missing. However, in the large-scale mask scenario, where we mask the lower right quarter of the area (resulting in 25\% data missing), the MAE and RMSE of UrbanFM reach 53.96 and 95.83, respectively, indicating worse performance despite a smaller proportion of missing data. This is because when the missing region is concentrated, the continuity and spatial correlation of the data are greatly affected, making it difficult for the model to leverage the effective information around the missing region for accurate prediction.

In addition, the performance of various methods can differ across different datasets, reflecting their unique characteristics. For example, the TaxiBJ and BikeNYC datasets, both traffic-related, exhibit more fluctuating data distributions, meaning values in nearby regions can vary significantly. In contrast, the TAPBJ dataset, focused on air quality, has a more uniform data distribution. With fixed masks, UrbanSTA shows a clear advantage over most comparison methods on the TaxiBJ and BikeNYC datasets, leveraging the specific dynamics of traffic data. However, on the TAPBJ dataset, it does not demonstrate a notable improvement, suggesting that its effectiveness may not extend to all types of data.

\subsection{Ablation study}

To thoroughly verify the effectiveness of each proposed module and our training strategy in DiffRecon, we compared its performance with various variants.

\begin{figure*}[tbp]
        \vspace{-10pt}
        \centering
        \subfloat[TaxiBJ-fix]
        {
            \includegraphics[width=0.23\linewidth]{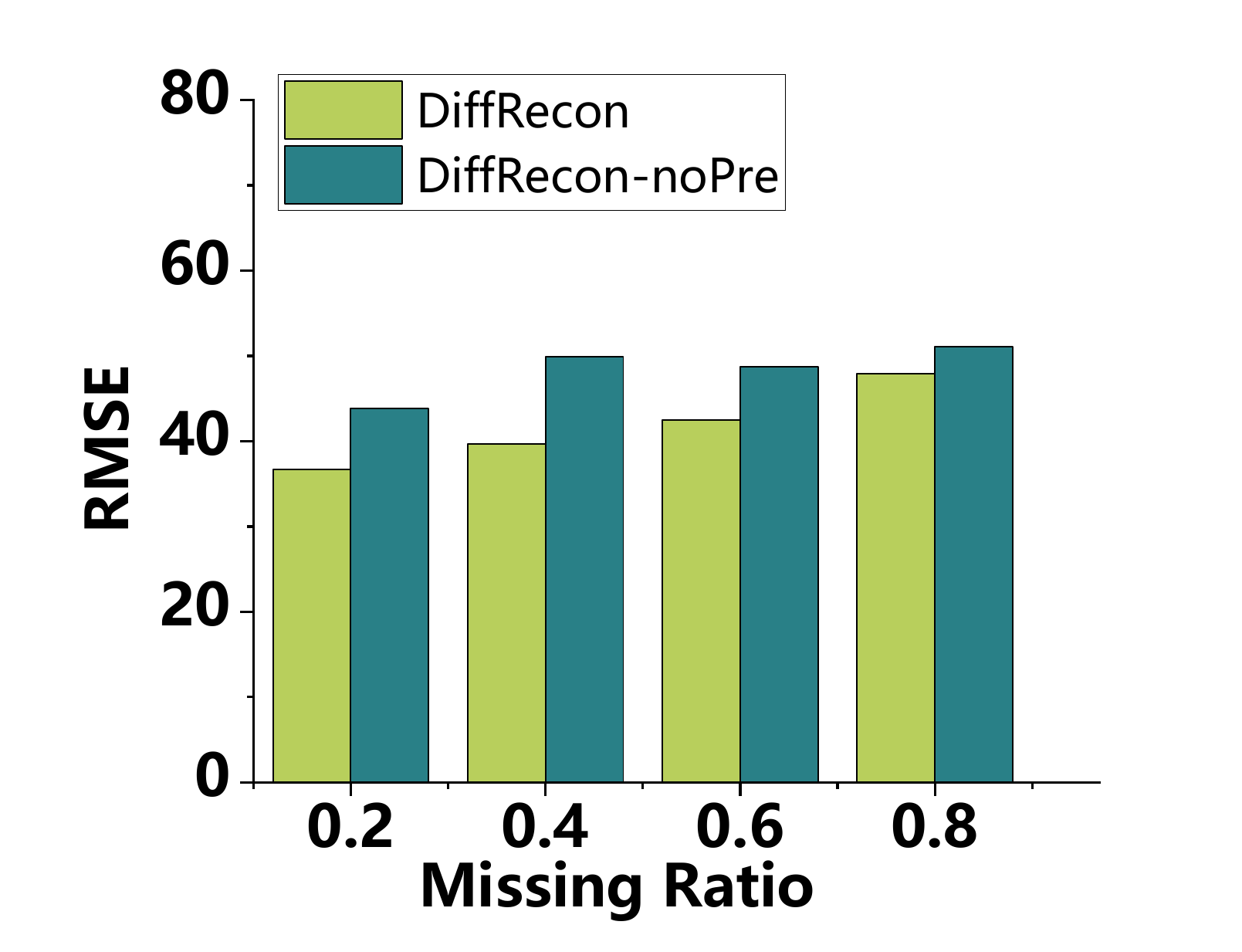}
        }
        \subfloat[TaxiBJ-random]
        {
            \includegraphics[width=0.23\linewidth]{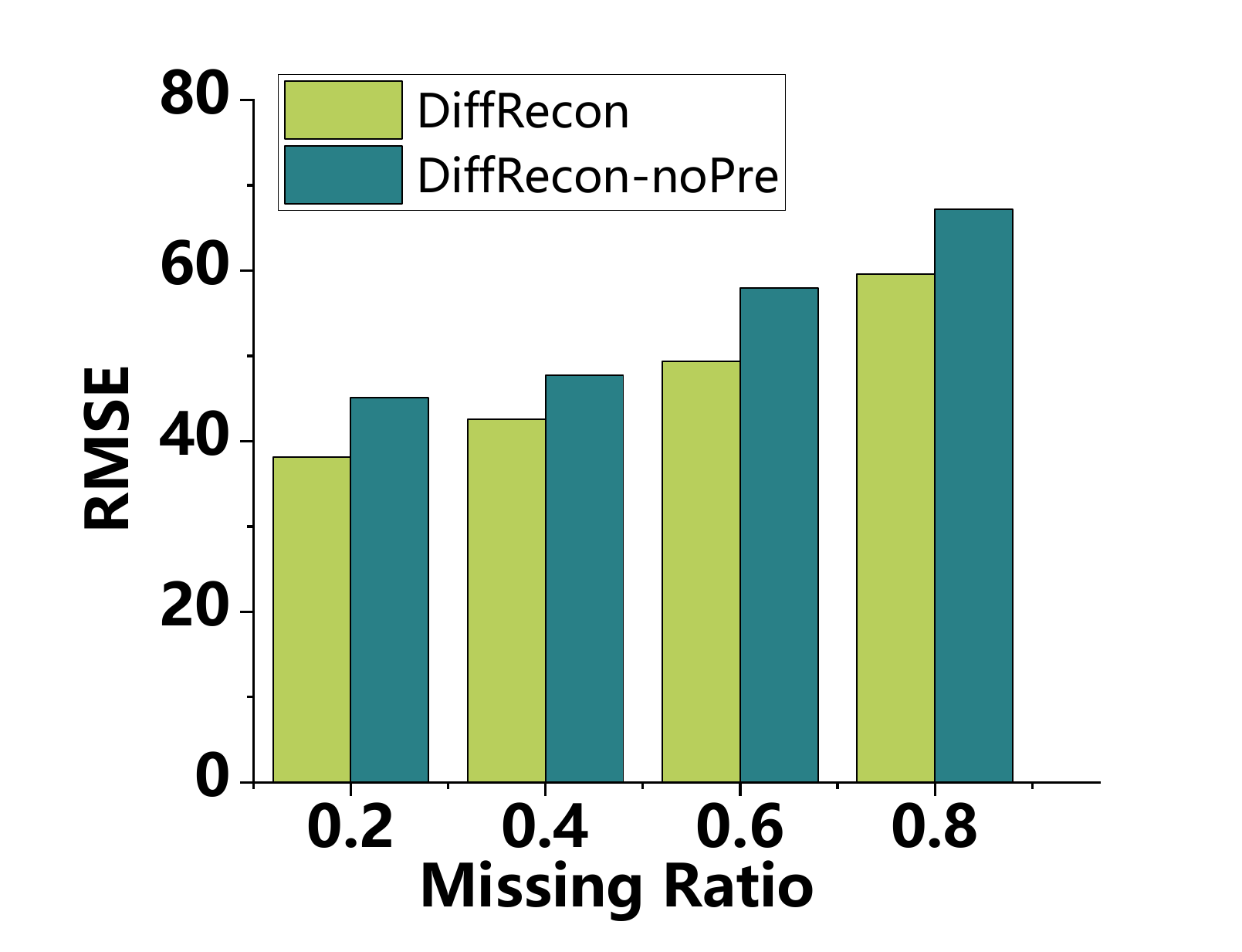}
        }
        \subfloat[BikeNYC-fix]
        {
            \includegraphics[width=0.23\linewidth]{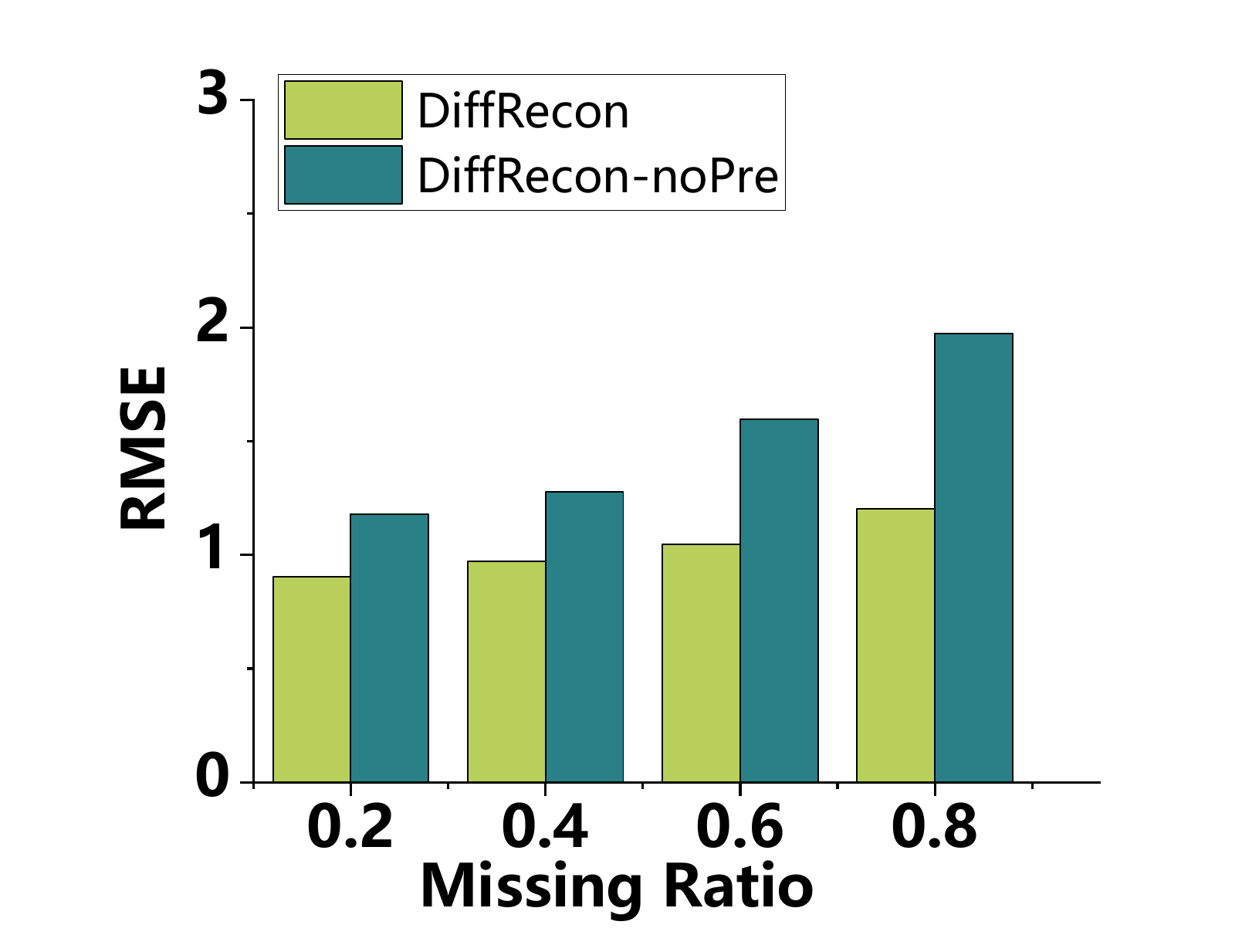}
        }
        \subfloat[BikeNYC-random]
        {
            \includegraphics[width=0.23\linewidth]{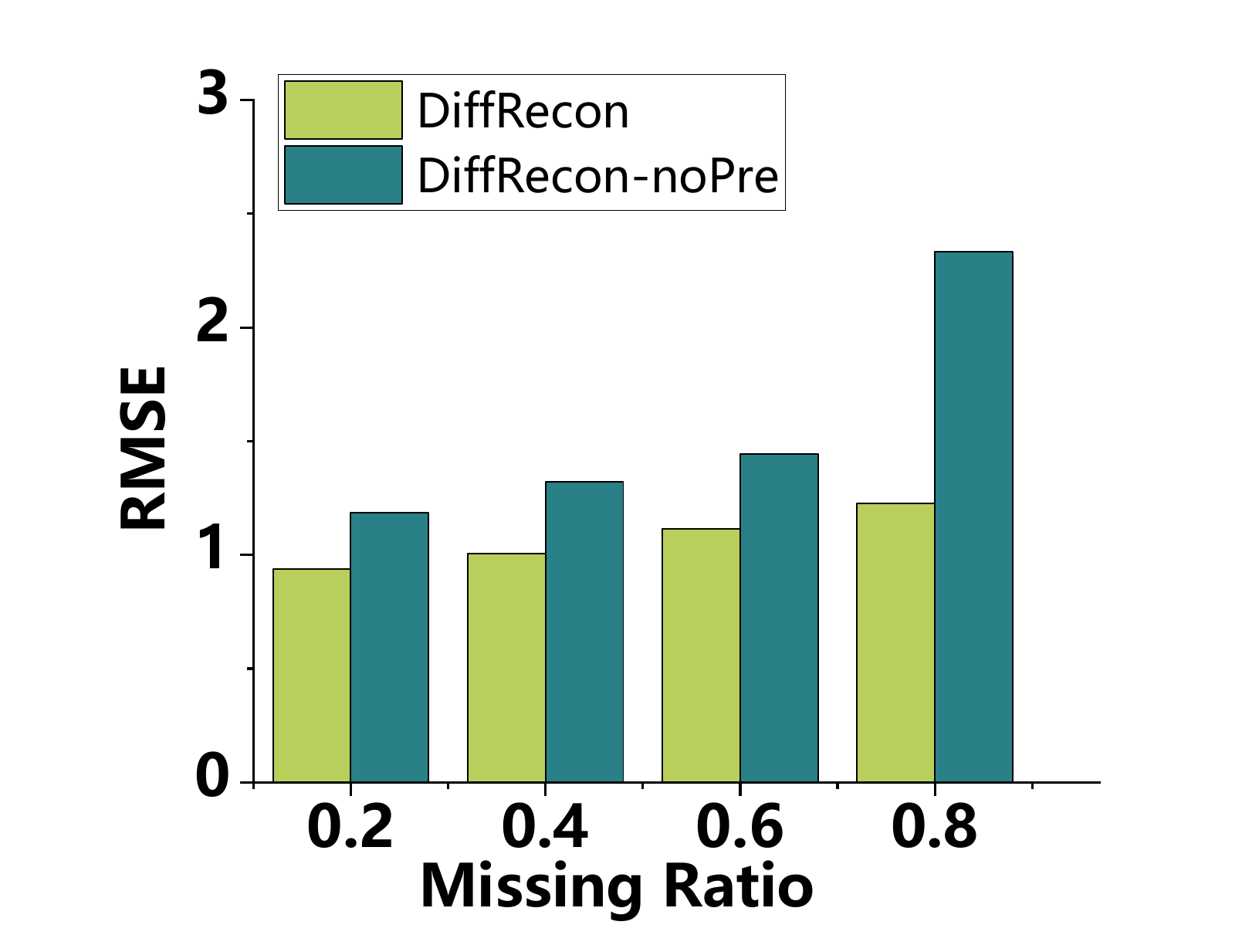}
        }
        \caption{Comparison of DiffRecon with DiffRecon without pre-training.}
        \label{fig:DiffRecon-noPre}   
        \vspace{-10pt}
\end{figure*}

\begin{figure*}[tbp]
        \vspace{-10pt}
        \centering
        \subfloat[TaxiBJ-fix]
        {
            \includegraphics[width=0.23\linewidth]{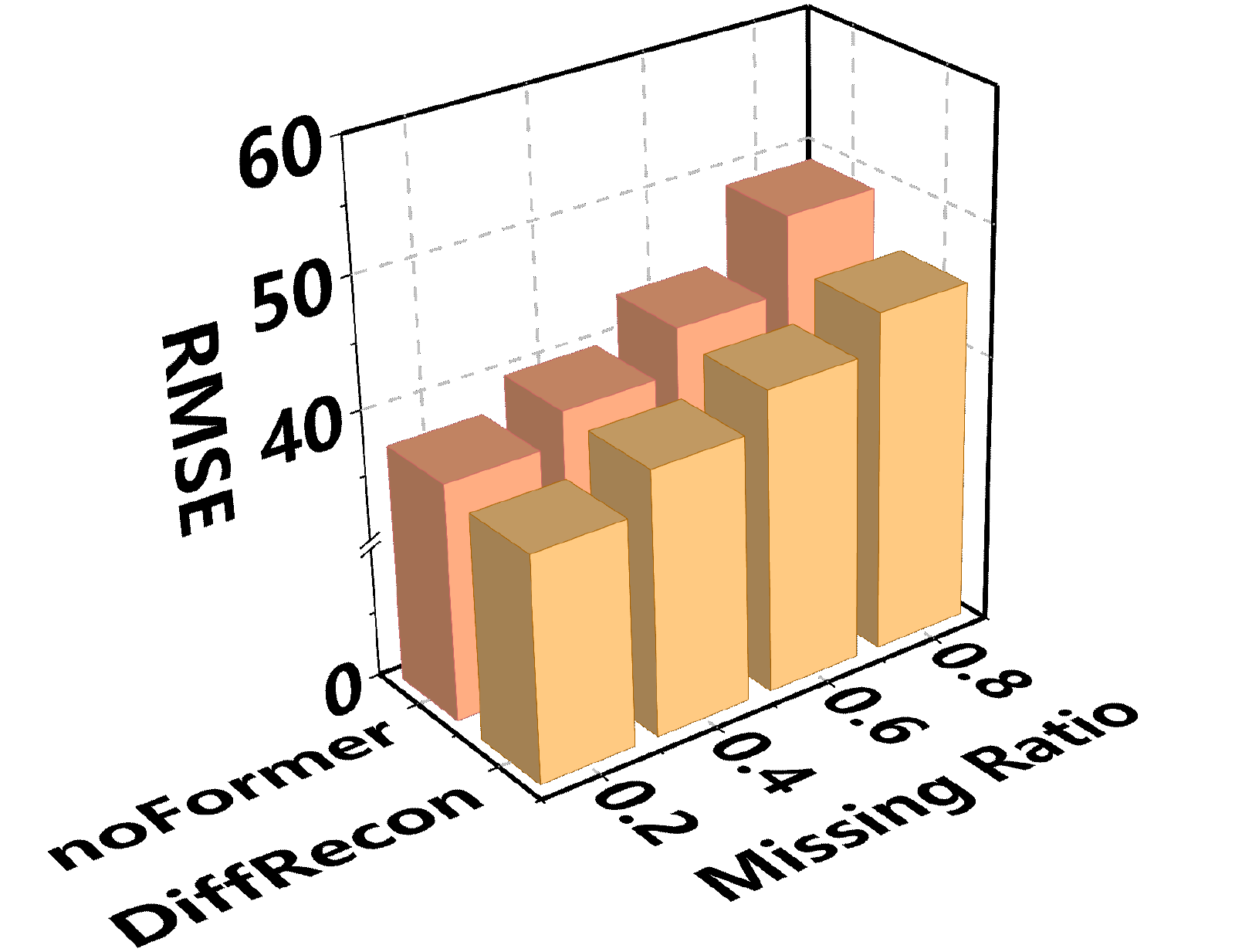}
        }
        \subfloat[TaxiBJ-random]
        {
            \includegraphics[width=0.23\linewidth]{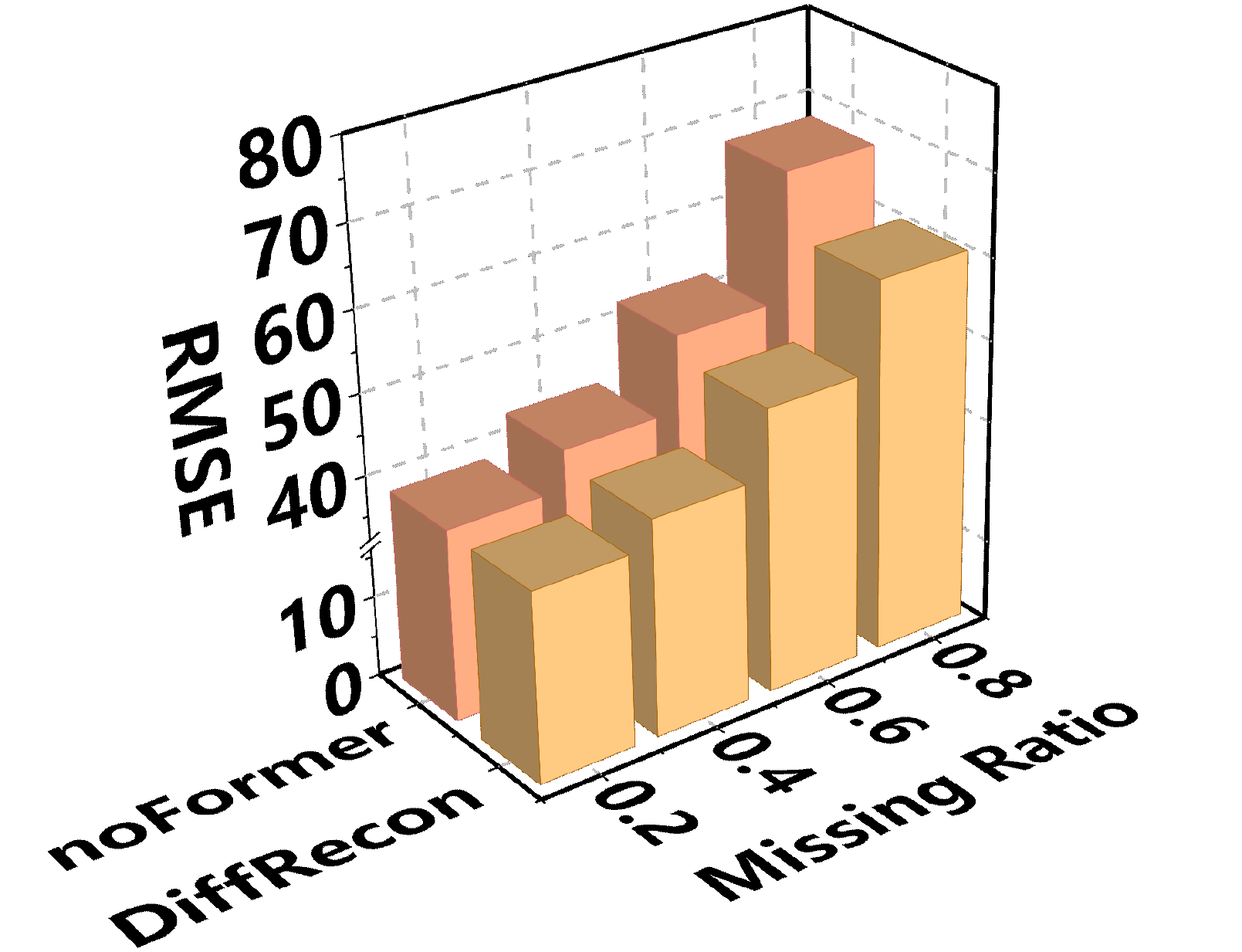}
        }
        \subfloat[TaxiBJ-fix]
        {
            \includegraphics[width=0.23\linewidth]{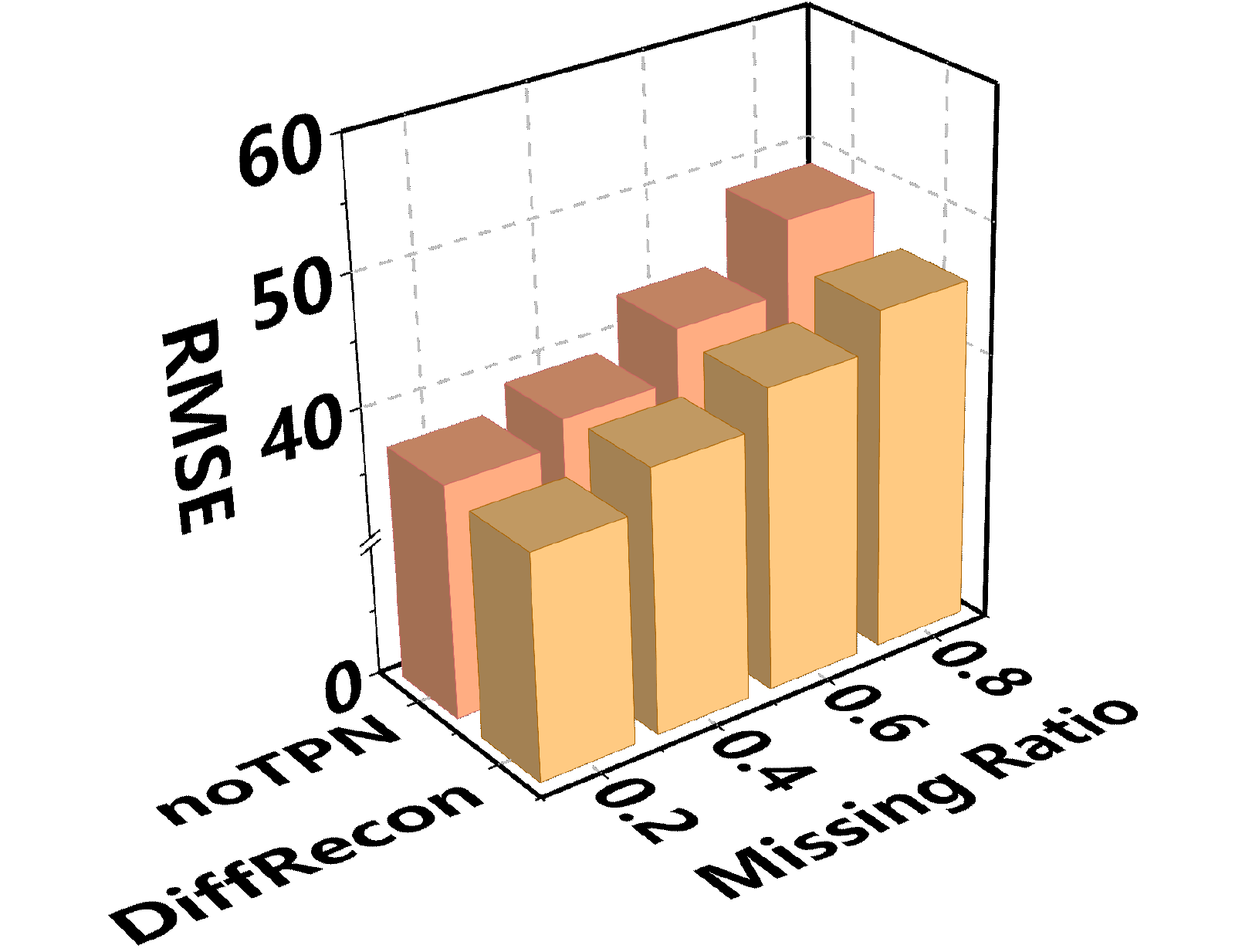}
        }
        \subfloat[TaxiBJ-random]
        {
            \includegraphics[width=0.23\linewidth]{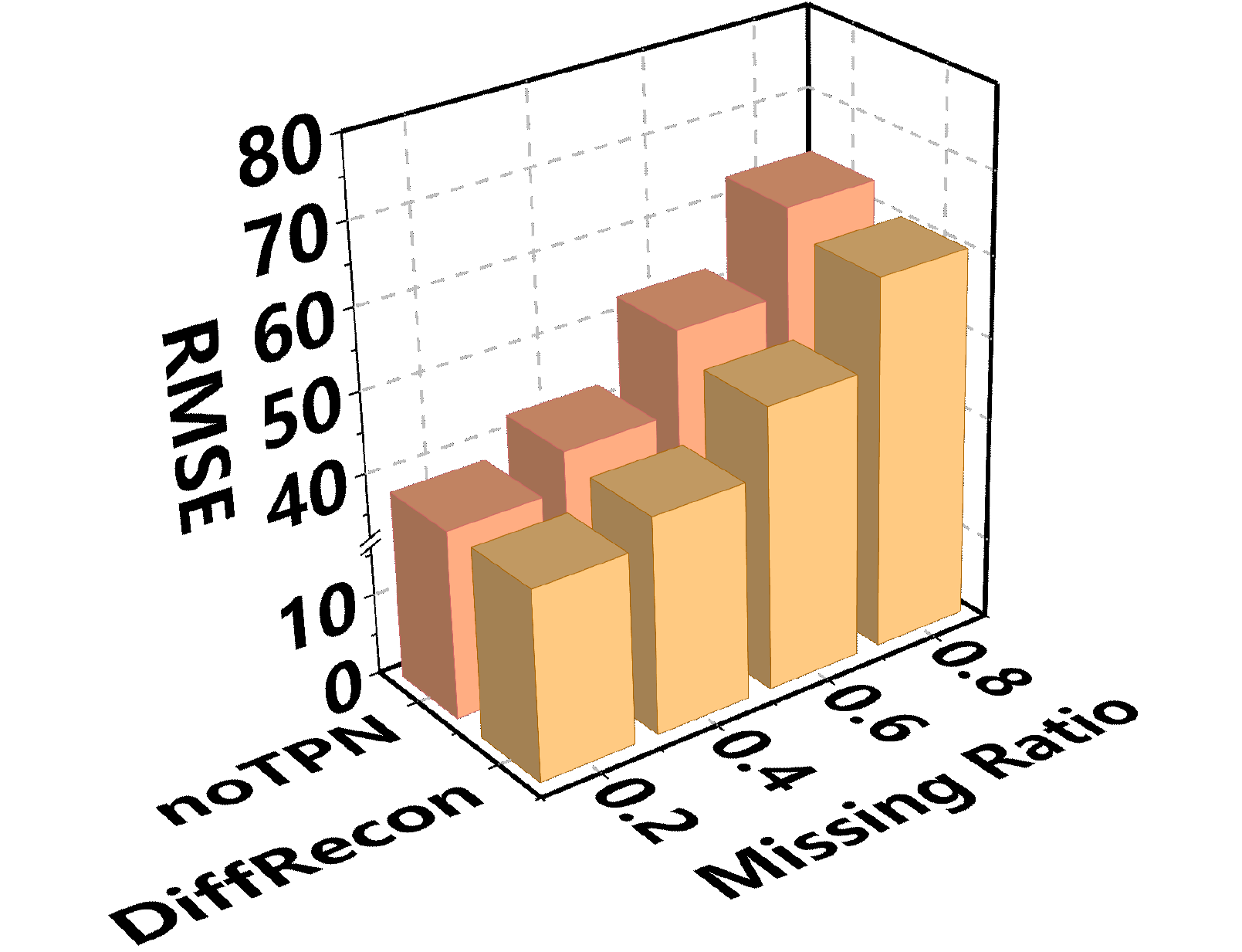}
        }
        \caption{Comparison of DiffRecon with DiffRecon without ST-PointFormer or T-PatternNet.}
        \label{fig:DiffRecon-noSTPointFormer/DiffRecon-noTPatternNet}   
        \vspace{-10pt}
\end{figure*}

\begin{figure}[tbp]
        \vspace{-10pt}
        \centering           
        \subfloat[TaxiBJ-fix]
        {
            \includegraphics[width=0.48\linewidth]{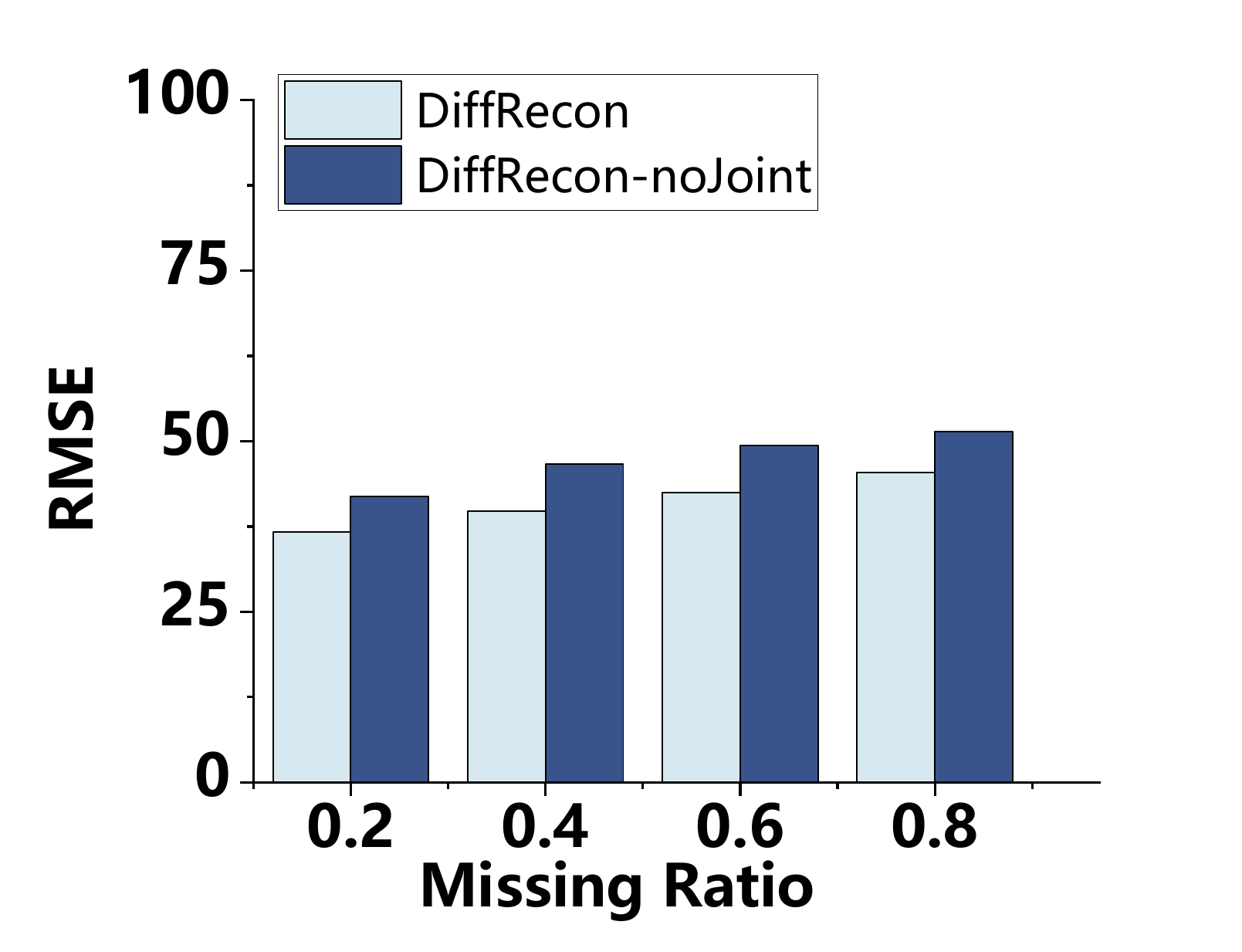}
        }
        \subfloat[TAPBJ-fix]
        {
            \includegraphics[width=0.48\linewidth]{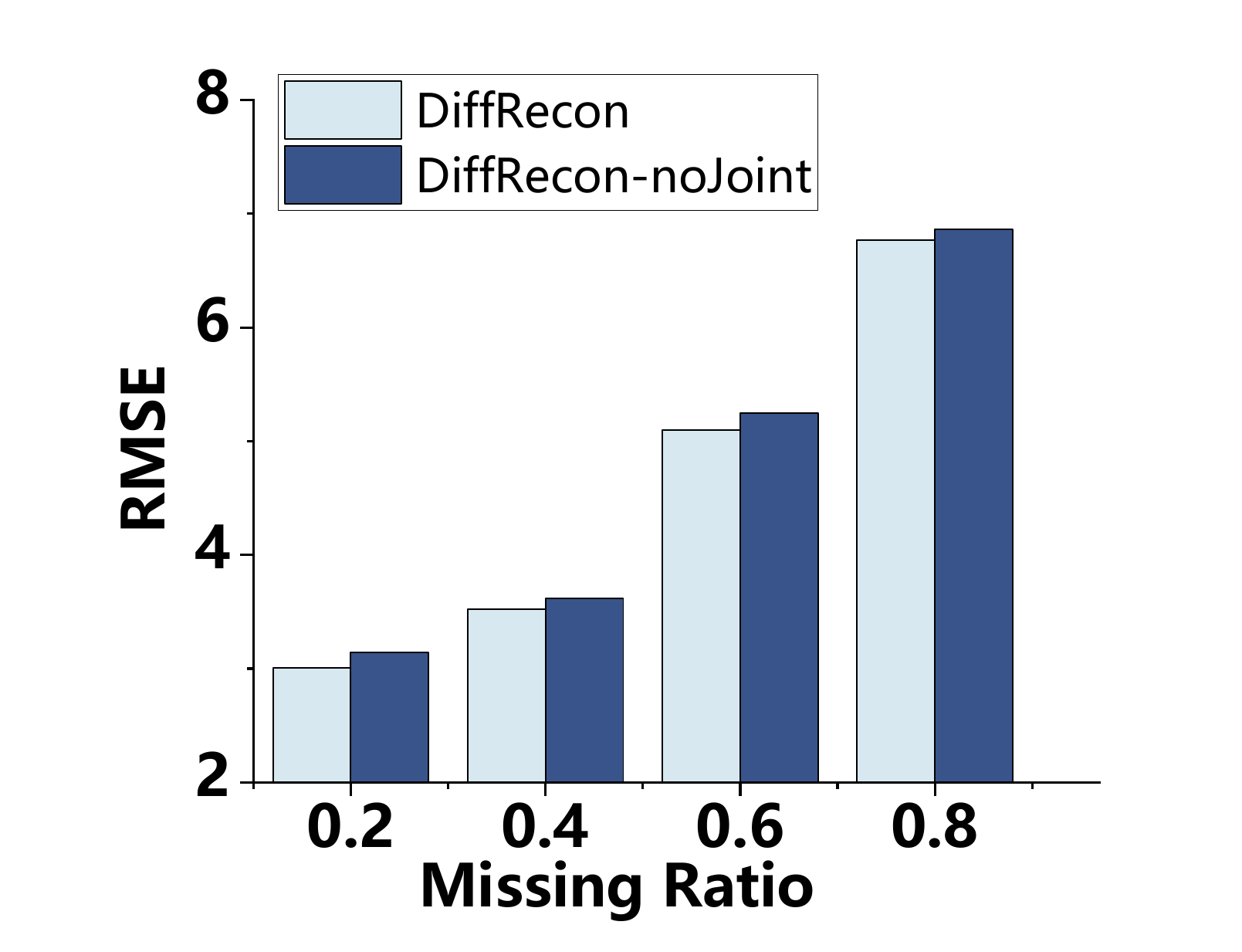}
        }
        \caption{Comparison of DiffRecon with DiffRecon without joint training.}
        \label{fig:DiffRecon-noJoint}   
        \vspace{-10pt}
\end{figure}

\subsubsection{DiffRecon-noPre}
    We did not perform pre-training but directly trained the model end-to-end for verification. From Figure~\ref{fig:DiffRecon-noPre}, we observe that DiffRecon-noPre performs significantly worse than DiffRecon across various sense ratios for both datasets, validating the necessity of the pre-training strategy. Additionally, the effect of the pre-training strategy is more pronounced in the BikeNYC dataset compared to the TaxiBJ dataset. This is because, although both datasets are traffic-related, TaxiBJ contains a clear and structured road network, while BikeNYC uses sparsely distributed stations as its basic statistical units. The data distribution in the latter is more uneven, making data reconstruction more challenging.

\subsubsection{DiffRecon-noJoint}
    We trained the two stages separately without joint training, using the output of the first stage as the input to the second stage and taking the output of the second stage as the final result to validate the importance of joint training fine-tuning. We conducted experiments on the fixed sensor task using the TaxiBJ and TAPBJ datasets. The results are shown in the Figure~\ref{fig:DiffRecon-noJoint}. When joint training is not applied, the model performance significantly declines, with a more pronounced effect on the TaxiBJ dataset. After removing joint training, the performance on the TaxiBJ dataset for 0.2, 0.4, 0.6, and 0.8 sense ratios dropped by 14.08\%, 17.61\%, 16.12\%, and 13.40\%, respectively. On the TAPBJ dataset, the performance dropped by 4.32\%, 2.84\%, 2.94\%, and 1.33\% for the same sense ratios.

\begin{figure}[tbp]
        \vspace{-10pt}
        \centering           
        \subfloat[TaxiBJ-fix]
        {
            \label{TaxiBJ-fix}\includegraphics[width=0.48\linewidth]{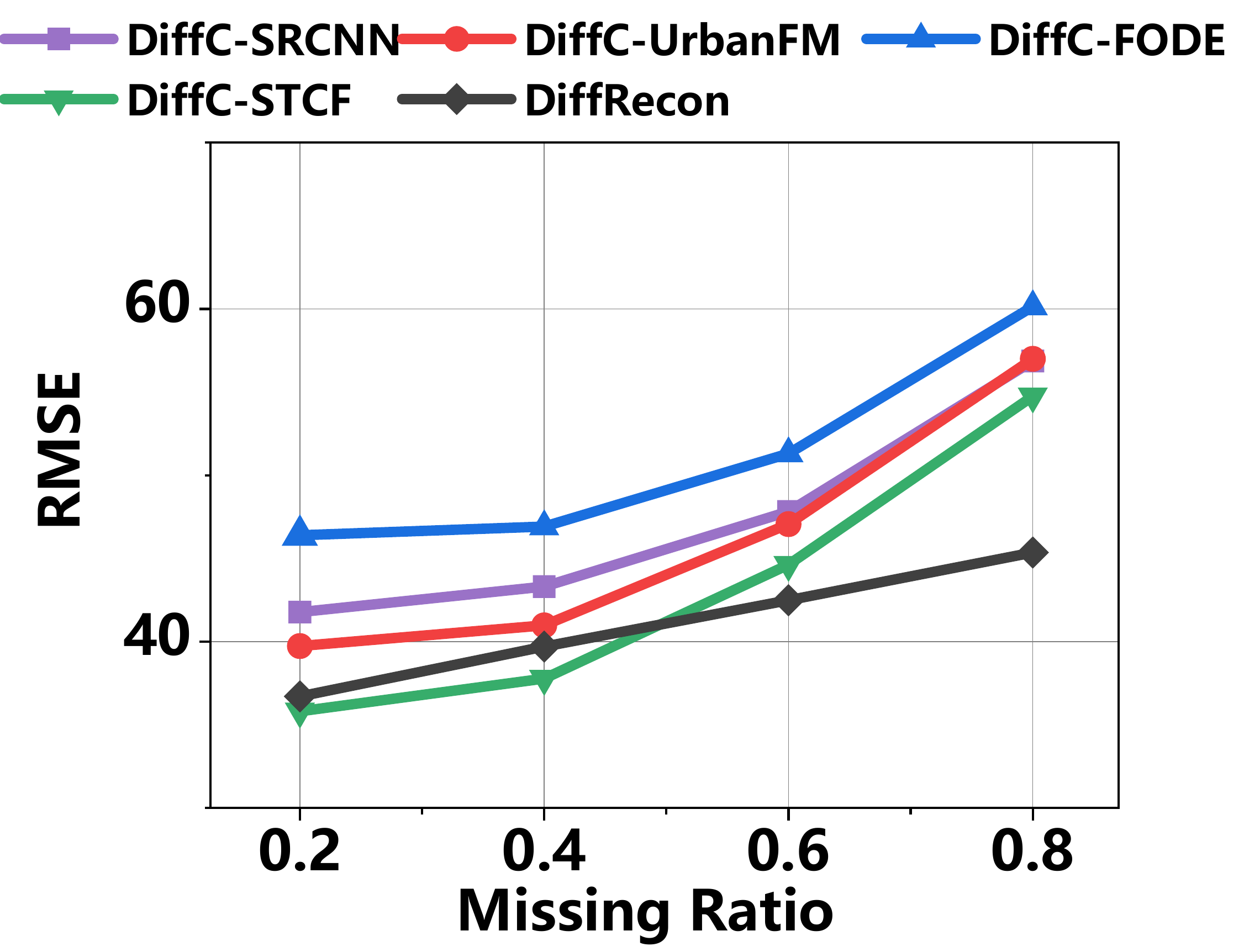}
        }
        \subfloat[TaxiBJ-random]
        {
            \label{TAPBJ-fix}\includegraphics[width=0.48\linewidth]{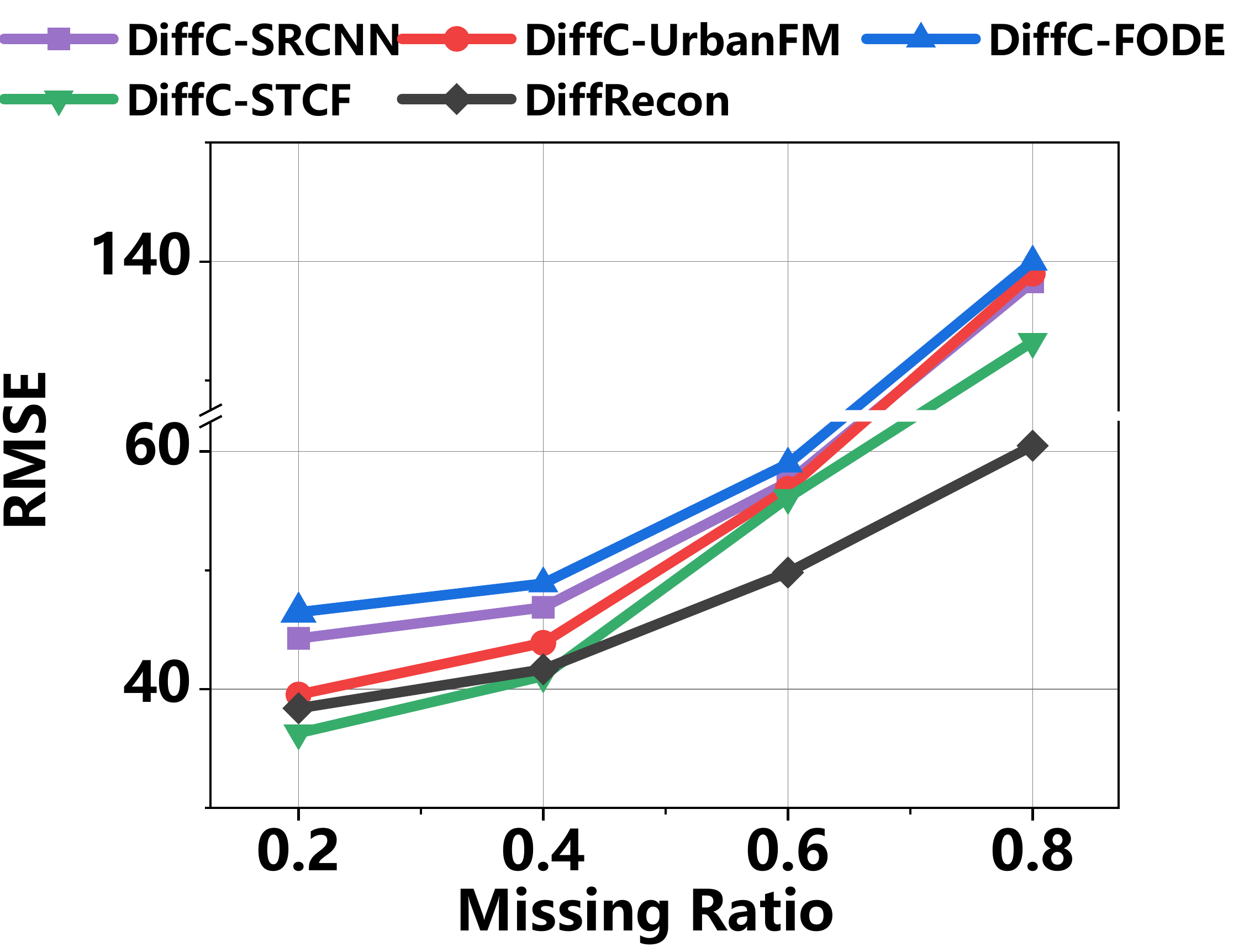}
        }
        \caption{Comparison of DiffRecon with DiffRecon that replaces Diffusion-F with existing fine-grained inference algorithms.}
        \label{fig:DiffRecon-noFG}   
        \vspace{-10pt}
\end{figure}

\subsubsection{DiffRecon-noSTPointFormer}
         We removed the ST-PointFormer in Stage 1, and the spatiotemporal characteristics of the data are not considered during the coarse-grained completion process. We conducted experiments on the TaxiBJ dataset, and the results are shown in the Figure~\ref{fig:DiffRecon-noSTPointFormer/DiffRecon-noTPatternNet}(a),(b). The performance of our model significantly decreased after removing the ST-PointFormer module, with the decline being most noticeable when the missing ratio was larger. When the missing ratios were 0.2 and 0.8, the performance dropped by 2.37\% and 8.16\%, respectively, after removing this module. This is because, when the missing ratio is high, the available information is limited, making it essential to fully leverage the spatiotemporal relationships among the sparse data to assist in data completion. The importance of the ST-PointFormer module becomes especially prominent in such scenarios.

\subsubsection{DiffRecon-noTPatternNet}
    We removed the T-PatternNet in Stage 2, and the temporal periodic relationships are not considered during the fine-grained inference process. The results are shown in the Figure~\ref{fig:DiffRecon-noSTPointFormer/DiffRecon-noTPatternNet}(c),(d). The experimental results are similar to those of DiffRecon-noSTPointFormer. When we removed the T-PatternNet module, the model's performance generally declined in most cases, with a more pronounced decrease when the missing ratio was higher.

\subsubsection{DiffRecon-noFG}
    Since most current research primarily focuses on inferring fine-grained spatiotemporal data from complete coarse-grained data, we replaced the fine-grained generation module in the second stage (Diffusion-F) with SRCNN, UrbanFM, FODE, and STCF to verify the irreplaceability of our designed fine-grained generation module in the fine-grained data reconstruction task. Similarly, we conducted experiments on the TaxiBJ dataset, and the experimental results are shown in the Figure~\ref{fig:DiffRecon-noFG}. After replacing Diffusion-F with SRCNN, UrbanFM, and FODE, the model's performance in various scenarios was consistently outperformed by DiffRecon. However, when replaced with the latest fine-grained spatiotemporal data inference method, STCF, it performed slightly better than DiffRecon under low missing ratios. As the missing ratio increased, the advantage of DiffRecon became more apparent. For example, in the fixed sensor scenario, when the missing ratio was 0.2, the performance of DiffRecon was 2.45\% lower than that of DiffC-STCF. However, when the missing ratios were 0.6 and 0.8, DiffRecon outperformed DiffC-STCF by 4.2\% and 17.17\%, respectively.

\begin{figure*}[!t]
\centerline{\includegraphics[width=1.05\linewidth]{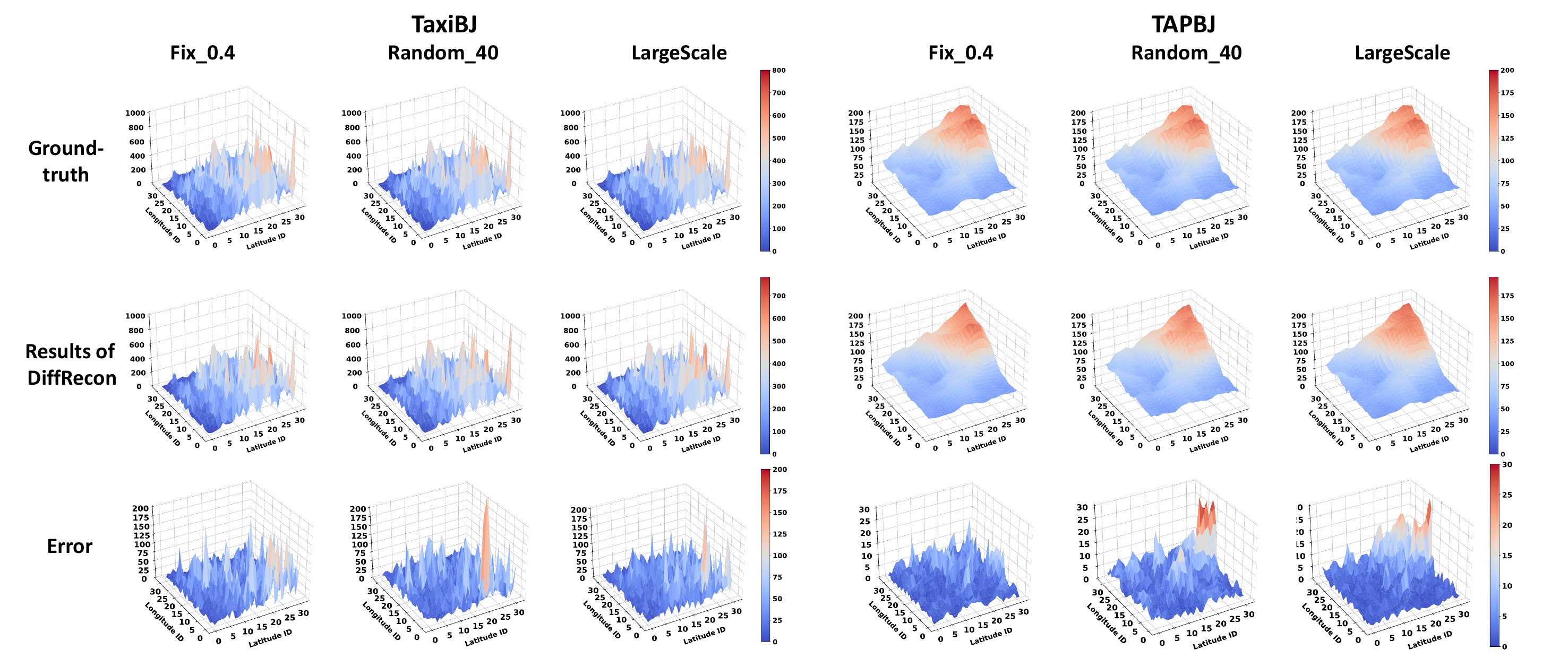}}
\caption{ 3D visualization heatmaps of three tasks and two datasets, TaxiBJ and TAPBJ, including ground truth, DiffRecon results and errors. }
\label{fig:3D Visualization}
\vspace{-15pt}
\end{figure*}

\subsection{Visualization}

Figure~\ref{fig:3D Visualization} presents some examples of DiffRecon's performance visualization. We use 3D heatmaps to display DiffRecon's generation results on the TaxiBJ and TAPBJ datasets across three task scenarios. These scenarios include: (1) each time slice with 40\% of the same fixed locations missing, (2) each time slice with 40\% of random locations missing, and (3) a quarter of the data missing in a large contiguous region. In the figure, the first row represents the ground truth. Since the time instances in the three task scenarios are the same, the ground truths for the three scenarios are identical for the TaxiBJ dataset, and the same applies to TAPBJ. The second row shows the generation results of DiffRecon in the various task scenarios. The third row illustrates the differences between the DiffRecon generation results and the ground truth, measured using Absolute Error.

It can be observed that the results generated by DiffRecon are very close to the ground truth, demonstrating the superiority of our method. Additionally, when comparing fixed missing data to random missing data, the fixed missing data shows smaller discrepancies from the ground truth, as evidenced by fewer or shorter peaks in the 3D error heatmap. Moreover, although the third task scenario (labeled "LargeScale" in the figure) has a missing rate of only 25\%, the large contiguous missing regions make it difficult to capture the spatial correlations between the missing locations and their surroundings. As a result, the performance of DiffRecon in this scenario is relatively worse, with the error map showing more and higher peaks. This is particularly evident in the TAPBJ dataset.

\section{Conclusion}
In this paper, we propose a novel task: spatiotemporal data reconstruction, which aims to infer complete, fine-grained data from incomplete and coarse-grained observations. To address this task, we introduce DiffRecon, a two-stage data inference framework. DiffRecon operates in a two-stage pipeline. In this stage, Diffusion-C performs coarse-grained completion and is enhanced by ST-PointFormer, an encoder that captures spatial relationships from sparse observations. This enhancement allows Diffusion-C to handle the lack of spatial information in sparse datasets, improving the quality of the reconstructed data. In the second stage, Diffusion-F handles fine-grained inference using a denoising network enhanced by T-PatternNet, a temporal encoder designed to extract complex patterns like periodicity and trends. Together, Diffusion-C and Diffusion-F form a robust framework for reconstructing fine-grained data. We validate the effectiveness of our approach through comparative experiments and ablation studies across multiple datasets, demonstrating the superiority of DiffRecon and its components in spatiotemporal data reconstruction.

\newpage

\bibliographystyle{IEEEtran}
\bibliography{reference}

\begin{thebibliography}{10}
\providecommand{\url}[1]{#1}
\csname url@samestyle\endcsname
\providecommand{\newblock}{\relax}
\providecommand{\bibinfo}[2]{#2}
\providecommand{\BIBentrySTDinterwordspacing}{\spaceskip=0pt\relax}
\providecommand{\BIBentryALTinterwordstretchfactor}{4}
\providecommand{\BIBentryALTinterwordspacing}{\spaceskip=\fontdimen2\font plus
\BIBentryALTinterwordstretchfactor\fontdimen3\font minus \fontdimen4\font\relax}
\providecommand{\BIBforeignlanguage}[2]{{%
\expandafter\ifx\csname l@#1\endcsname\relax
\typeout{** WARNING: IEEEtran.bst: No hyphenation pattern has been}%
\typeout{** loaded for the language `#1'. Using the pattern for}%
\typeout{** the default language instead.}%
\else
\language=\csname l@#1\endcsname
\fi
#2}}
\providecommand{\BIBdecl}{\relax}
\BIBdecl

\bibitem{han2021joint}
J.~Han, H.~Liu, H.~Zhu, H.~Xiong, and D.~Dou, ``Joint air quality and weather prediction based on multi-adversarial spatiotemporal networks,'' in \emph{Proceedings of the AAAI Conference on Artificial Intelligence}, vol.~35, no.~5, May 2021, pp. 4081--4089.

\bibitem{yuan2021survey}
H.~Yuan and G.~Li, ``A survey of traffic prediction: from spatio-temporal data to intelligent transportation,'' \emph{Data Science and Engineering}, vol.~6, no.~1, pp. 63--85, 2021.

\bibitem{liu2022practical}
F.~Liu, H.~Liu, and W.~Jiang, ``Practical adversarial attacks on spatiotemporal traffic forecasting models,'' \emph{Advances in Neural Information Processing Systems}, vol.~35, pp. 19\,035--19\,047, 2022.

\bibitem{cheng2020short}
S.~Cheng, F.~Lu, and P.~Peng, ``Short-term traffic forecasting by mining the non-stationarity of spatiotemporal patterns,'' \emph{IEEE Transactions on Intelligent Transportation Systems}, vol.~22, no.~10, pp. 6365--6383, 2020.

\bibitem{meng2021cross}
C.~Meng, S.~Rambhatla, and Y.~Liu, ``Cross-node federated graph neural network for spatio-temporal data modeling,'' in \emph{Proceedings of the 27th ACM SIGKDD conference on knowledge discovery \& data mining}, 2021, pp. 1202--1211.

\bibitem{wang2020survey}
S.~Wang, J.~Cao, and S.~Y. Philip, ``Deep learning for spatio-temporal data mining: A survey,'' \emph{IEEE Transactions on Knowledge and Data Engineering}, vol.~34, no.~8, pp. 3681--3700, 2020.

\bibitem{ouyang2020fine}
K.~Ouyang, Y.~Liang, Y.~Liu, Z.~Tong, S.~Ruan, Y.~Zheng, and D.~S. Rosenblum, ``Fine-grained urban flow inference,'' \emph{IEEE Transactions on Knowledge and Data Engineering}, vol.~34, no.~6, pp. 2755--2770, 2020.

\bibitem{liu2023pristi}
M.~Liu, H.~Huang, H.~Feng, L.~Sun, B.~Du, and Y.~Fu, ``Pristi: A conditional diffusion framework for spatiotemporal imputation,'' in \emph{2023 IEEE 39th International Conference on Data Engineering (ICDE)}.\hskip 1em plus 0.5em minus 0.4em\relax IEEE, April 2023, pp. 1927--1939.

\bibitem{liang2019urbanfm}
Y.~Liang, K.~Ouyang, L.~Jing, S.~Ruan, Y.~Liu, J.~Zhang, and Y.~Zheng, ``Urbanfm: Inferring fine-grained urban flows,'' in \emph{Proceedings of the 25th ACM SIGKDD International Conference on Knowledge Discovery \& Data Mining}.\hskip 1em plus 0.5em minus 0.4em\relax ACM, July 2019, pp. 3132--3142.

\bibitem{xu2023diffusion}
X.~Xu, Y.~Wei, P.~Wang, X.~Luo, F.~Zhou, and G.~Trajcevski, ``Diffusion probabilistic modeling for fine-grained urban traffic flow inference with relaxed structural constraint,'' in \emph{ICASSP 2023-2023 IEEE International Conference on Acoustics, Speech and Signal Processing (ICASSP)}.\hskip 1em plus 0.5em minus 0.4em\relax IEEE, June 2023, pp. 1--5.

\bibitem{zhou2020enhancing}
F.~Zhou, L.~Li, T.~Zhong, G.~Trajcevski, K.~Zhang, and J.~Wang, ``Enhancing urban flow maps via neural odes,'' in \emph{Proceedings of the Twenty-Ninth International Joint Conference on Artificial Intelligence,$\{$IJCAI$\}$ 2020}, 2020.

\bibitem{wang2020calendar}
D.~Wang, M.~Jiang, M.~Syed, O.~Conway, V.~Juneja, S.~Subramanian, and N.~V. Chawla, ``Calendar graph neural networks for modeling time structures in spatiotemporal user behaviors,'' in \emph{Proceedings of the 26th ACM SIGKDD international conference on knowledge discovery \& data mining}, 2020, pp. 2581--2589.

\bibitem{wang2023urbanSTA}
R.~Wang, Y.~Liu, Y.~Gong, W.~Liu, M.~Chen, Y.~Yin, and Y.~Zheng, ``Fine-grained urban flow inference with unobservable data via space-time attraction learning,'' in \emph{2023 IEEE International Conference on Data Mining (ICDM)}.\hskip 1em plus 0.5em minus 0.4em\relax IEEE, December 2023, pp. 1367--1372.

\bibitem{ho2020denoising}
J.~Ho, A.~Jain, and P.~Abbeel, ``Denoising diffusion probabilistic models,'' \emph{Advances in neural information processing systems}, vol.~33, pp. 6840--6851, 2020.

\bibitem{pathak2016context}
D.~Pathak, P.~Krahenbuhl, J.~Donahue, T.~Darrell, and A.~A. Efros, ``Context encoders: Feature learning by inpainting,'' in \emph{Proceedings of the IEEE conference on computer vision and pattern recognition}, 2016, pp. 2536--2544.

\bibitem{yu2018generative}
J.~Yu, Z.~Lin, J.~Yang, X.~Shen, X.~Lu, and T.~S. Huang, ``Generative image inpainting with contextual attention,'' in \emph{Proceedings of the IEEE conference on computer vision and pattern recognition}, 2018, pp. 5505--5514.

\bibitem{wu2021spatial}
Y.~Wu, D.~Zhuang, M.~Lei, A.~Labbe, and L.~Sun, ``Spatial aggregation and temporal convolution networks for real-time kriging,'' \emph{arXiv preprint arXiv:2109.12144}, 2021.

\bibitem{xu2019fine}
Y.~Xu, Y.~Zhu, Y.~Shen, and J.~Yu, ``Fine-grained air quality inference with remote sensing data and ubiquitous urban data,'' \emph{ACM Transactions on Knowledge Discovery from Data (TKDD)}, vol.~13, no.~5, pp. 1--27, 2019.

\bibitem{wang2020deep}
E.~Wang, M.~Zhang, X.~Cheng, Y.~Yang, W.~Liu, H.~Yu, L.~Wang, and J.~Zhang, ``Deep learning-enabled sparse industrial crowdsensing and prediction,'' \emph{IEEE Transactions on Industrial Informatics}, vol.~17, no.~9, pp. 6170--6181, 2020.

\bibitem{wang2023spatiotemporal}
E.~Wang, W.~Liu, W.~Liu, C.~Xiang, B.~Yang, and Y.~Yang, ``Spatiotemporal transformer for data inference and long prediction in sparse mobile crowdsensing,'' in \emph{IEEE INFOCOM 2023-IEEE Conference on Computer Communications}.\hskip 1em plus 0.5em minus 0.4em\relax IEEE, May 2023, pp. 1--10.

\bibitem{dong2015image}
C.~Dong, C.~C. Loy, K.~He, and X.~Tang, ``Image super-resolution using deep convolutional networks,'' \emph{IEEE transactions on pattern analysis and machine intelligence}, vol.~38, no.~2, pp. 295--307, 2015.

\bibitem{ledig2017photo}
C.~Ledig, L.~Theis, F.~Husz{\'a}r, J.~Caballero, A.~Cunningham, A.~Acosta, A.~Aitken, A.~Tejani, J.~Totz, Z.~Wang \emph{et~al.}, ``Photo-realistic single image super-resolution using a generative adversarial network,'' in \emph{Proceedings of the IEEE conference on computer vision and pattern recognition}, 2017, pp. 4681--4690.

\bibitem{xu2023spatial}
X.~Xu, Z.~Wang, Q.~Gao, T.~Zhong, B.~Hui, F.~Zhou, and G.~Trajcevski, ``Spatial-temporal contrasting for fine-grained urban flow inference,'' \emph{IEEE Transactions on Big Data}, 2023.

\bibitem{li2022fine}
J.~Li, S.~Wang, J.~Zhang, H.~Miao, J.~Zhang, and S.~Y. Philip, ``Fine-grained urban flow inference with incomplete data,'' \emph{IEEE Transactions on Knowledge and Data Engineering}, vol.~35, no.~6, pp. 5851--5864, 2022.

\bibitem{zheng2023diffuflow}
Y.~Zheng, L.~Zhong, S.~Wang, Y.~Yang, W.~Gu, J.~Zhang, and J.~Wang, ``Diffuflow: Robust fine-grained urban flow inference with denoising diffusion model,'' in \emph{Proceedings of the 32nd ACM International Conference on Information and Knowledge Management}, 2023, pp. 3505--3513.

\bibitem{rombach2022high}
R.~Rombach, A.~Blattmann, D.~Lorenz, P.~Esser, and B.~Ommer, ``High-resolution image synthesis with latent diffusion models,'' in \emph{Proceedings of the IEEE/CVF conference on computer vision and pattern recognition}, 2022, pp. 10\,684--10\,695.

\bibitem{croitoru2023diffusion}
F.-A. Croitoru, V.~Hondru, R.~T. Ionescu, and M.~Shah, ``Diffusion models in vision: A survey,'' \emph{IEEE Transactions on Pattern Analysis and Machine Intelligence}, vol.~45, no.~9, pp. 10\,850--10\,869, 2023.

\bibitem{li2022diffusion}
X.~Li, J.~Thickstun, I.~Gulrajani, P.~S. Liang, and T.~B. Hashimoto, ``Diffusion-lm improves controllable text generation,'' \emph{Advances in Neural Information Processing Systems}, vol.~35, pp. 4328--4343, 2022.

\bibitem{gong2022diffuseq}
S.~Gong, M.~Li, J.~Feng, Z.~Wu, and L.~Kong, ``Diffuseq: Sequence to sequence text generation with diffusion models,'' \emph{arXiv preprint arXiv:2210.08933}, 2022.

\bibitem{gu2022vector}
S.~Gu, D.~Chen, J.~Bao, F.~Wen, B.~Zhang, D.~Chen, L.~Yuan, and B.~Guo, ``Vector quantized diffusion model for text-to-image synthesis,'' in \emph{Proceedings of the IEEE/CVF conference on computer vision and pattern recognition}, 2022, pp. 10\,696--10\,706.

\bibitem{bao2023one}
F.~Bao, S.~Nie, K.~Xue, C.~Li, S.~Pu, Y.~Wang, G.~Yue, Y.~Cao, H.~Su, and J.~Zhu, ``One transformer fits all distributions in multi-modal diffusion at scale,'' in \emph{International Conference on Machine Learning}.\hskip 1em plus 0.5em minus 0.4em\relax PMLR, 2023, pp. 1692--1717.

\bibitem{tashiro2021csdi}
Y.~Tashiro, J.~Song, Y.~Song, and S.~Ermon, ``Csdi: Conditional score-based diffusion models for probabilistic time series imputation,'' \emph{Advances in Neural Information Processing Systems}, vol.~34, pp. 24\,804--24\,816, 2021.

\bibitem{kong2020diffwave}
Z.~Kong, W.~Ping, J.~Huang, K.~Zhao, and B.~Catanzaro, ``Diffwave: A versatile diffusion model for audio synthesis,'' \emph{arXiv preprint arXiv:2009.09761}, 2020.

\bibitem{lidiffstitch}
G.~Li, Y.~Shan, Z.~Zhu, T.~Long, and W.~Zhang, ``Diffstitch: Boosting offline reinforcement learning with diffusion-based trajectory stitching,'' in \emph{Forty-first International Conference on Machine Learning}.

\bibitem{kang2024efficient}
B.~Kang, X.~Ma, C.~Du, T.~Pang, and S.~Yan, ``Efficient diffusion policies for offline reinforcement learning,'' \emph{Advances in Neural Information Processing Systems}, vol.~36, 2024.

\bibitem{nie2022diffusion}
W.~Nie, B.~Guo, Y.~Huang, C.~Xiao, A.~Vahdat, and A.~Anandkumar, ``Diffusion models for adversarial purification,'' \emph{arXiv preprint arXiv:2205.07460}, 2022.

\bibitem{sun2023enhance}
J.~Sun, S.~Sinha, and A.~Zhang, ``Enhance diffusion to improve robust generalization,'' in \emph{Proceedings of the 29th ACM SIGKDD Conference on Knowledge Discovery and Data Mining}, 2023, pp. 2083--2095.

\bibitem{wu2022timesnet}
H.~Wu, T.~Hu, Y.~Liu, H.~Zhou, J.~Wang, and M.~Long, ``Timesnet: Temporal 2d-variation modeling for general time series analysis,'' \emph{arXiv preprint arXiv:2210.02186}, 2022.

\bibitem{zhang2017deep}
J.~Zhang, Y.~Zheng, and D.~Qi, ``Deep spatio-temporal residual networks for citywide crowd flows prediction,'' in \emph{Proceedings of the AAAI conference on artificial intelligence}, vol.~31, no.~1, 2017.

\bibitem{geng2021tracking}
G.~Geng, Q.~Xiao, S.~Liu, X.~Liu, J.~Cheng, Y.~Zheng, T.~Xue, D.~Tong, B.~Zheng, Y.~Peng \emph{et~al.}, ``Tracking air pollution in china: near real-time pm2. 5 retrievals from multisource data fusion,'' \emph{Environmental Science \& Technology}, vol.~55, no.~17, pp. 12\,106--12\,115, 2021.

\bibitem{xiao2022spatiotemporal}
Q.~Xiao, G.~Geng, S.~Liu, J.~Liu, X.~Meng, and Q.~Zhang, ``Spatiotemporal continuous estimates of daily 1 km pm 2.5 from 2000 to present under the tracking air pollution in china (tap) framework,'' \emph{Atmospheric Chemistry and Physics}, vol.~22, no.~19, pp. 13\,229--13\,242, 2022.

\bibitem{xiao2021separating}
Q.~Xiao, Y.~Zheng, G.~Geng, C.~Chen, X.~Huang, H.~Che, X.~Zhang, K.~He, and Q.~Zhang, ``Separating emission and meteorological contribution to pm 2.5 trends over east china during 2000--2018,'' \emph{Atmospheric Chemistry and Physics Discussions}, vol. 2021, pp. 1--32, 2021.

\end{thebibliography}

\end{document}